\documentclass{article}
\usepackage[size=normalsize,font=sf,labelfont=bf]{caption}

\usepackage{geometry}
\usepackage{subcaption}
\usepackage{graphicx}
\usepackage{authblk}
\usepackage{hyperref}
\usepackage{threeparttable}
\usepackage{fancyhdr}
\usepackage{datetime} % for auto datetime
\usepackage[nodisplayskipstretch]{setspace} % https://tex.stackexchange.com/questions/32918/how-can-i-decrease-spaces-between-equations

\usepackage{bm} % bold match, e.g. for matrix and vector
\usepackage{makecell} % line breaks in table cells https://tex.stackexchange.com/questions/2441/how-to-add-a-forced-line-break-inside-a-table-cell/19678
\usepackage{dblfloatfix}    % To enable figures at the bottom of page
\usepackage{multirow} % https://ctan.org/pkg/multirow?lang=en
\usepackage{booktabs} % https://ctan.org/pkg/booktabs?lang=en 
\usepackage{diagbox}
\usepackage{amssymb}
\usepackage[abbreviations]{foreign} % https://tex.stackexchange.com/questions/15009/macros-for-common-abbreviations
\usepackage{hyperref}
\usepackage{float}
\usepackage{textcomp} % \textdegree symbol
\usepackage[style=nature, autocite    = superscript, backend=biber, defernumbers=true,maxnames=6]{biblatex} % https://tex.stackexchange.com/questions/424774/print-url-only-if-doi-not-present
\DeclareSourcemap{
  \maps[datatype=bibtex]{
    \map[overwrite]{
      \step[fieldsource=doi, final]
      \step[fieldset=url, null]
      \step[fieldset=eprint, null]
    }  
  }
}
\AtEveryBibitem{%
  \clearlist{language}% https://tex.stackexchange.com/questions/121661/biblatex-remove-the-eng-from-citations
}
\makeatother

\makeatletter
\newcommand*{\addFileDependency}[1]{% argument=file name and extension
  \typeout{(#1)}% latexmk will find this if $recorder=0 (however, in that case, it will ignore #1 if it is a .aux or .pdf file etc and it exists! if it doesn't exist, it will appear in the list of dependents regardless)
  \@addtofilelist{#1}% if you want it to appear in \listfiles, not really necessary and latexmk doesn't use this
  \IfFileExists{#1}{}{\typeout{No file #1.}}% latexmk will find this message if #1 doesn't exist (yet)
}
\makeatother

\usepackage[dvipsnames]{xcolor} % used for colored acronyms

\usepackage[draft]{pdfcomment} % https://ctan.org/pkg/pdfcomment?lang=en 
\usepackage[hyperfirst=false,acronym,sort=none,shortcuts,nopostdot,style=super,nonumberlist,toc,nogroupskip]{glossaries}

\glsdisablehyper % to disable hyperlinks on all acronyms

\newcommand{\accolor}[1]{\textcolor{Sepia}{#1}}
\newacronymstyle{myacro}
{%
  \GlsUseAcrEntryDispStyle{long-short}%
}%
{%
  \GlsUseAcrStyleDefs{long-short}%
}

\setacronymstyle{myacro}

\newcommand*{\tip}[1]{%  define our acronym command,  make it short since we use it a lot, use * for so that it is only a 'short' command
    \ifglsused{#1}{% if we used it already, then put pdftooltip
      {\pdftooltip{\accolor{\glsentryshort{#1}}}{\glsentrydesc{#1}}}%
    }{%
      \gls{#1}% otherwise put the normal gls
    }%
}%

\newcommand*{\tips}[1]{%  define our acronym command,  make it short since we use it a lot, use * for so that it is only a 'short' command
    \ifglsused{#1}{% if we used it already, then put pdftooltip
      {\pdftooltip{\accolor{\glsentryshortpl{#1}}}{\glsentrydescplural{#1}}}%
    }{%
      \glspl{#1}% otherwise put the normal gls
    }%
}%

\newcommand*{\tipshort}[1]{%  define our acronym command,  make it short since we use it a lot, use * for so that it is only a 'short' command
      {\pdftooltip{\accolor{\glsentryshort{#1}}}{\glsentrydesc{#1}}}%
}%

\newacronym{adc}{ADC}{Analog to Digital Converter}
\newacronym[description={A candidate or best-match Target Block for block matching in the t-d2 slice}]{tb}{TB}{Target Block}
\newacronym[description={A Reference Block centered on the event location in the t-d1 slice}]{rb}{RB}{Reference Block}
\newacronym[description={Coarse to Fine hierarchical search strategy used in BMOF}]{ctf}{CTF}{Coarse to Fine}
\newacronym[description={Coefficient Of Variation (sigma/mean)}]{cov}{COV}{Coefficient Of Variation}
\newacronym[description={Corner point in an image}]{kp}{KP}{Keypoint}
\newacronym[description={Direction Selective model of motion detection in biological vision, usually either Hassenstein-Reichhardt or Barlow-Levick type}]{ds}{DS}{Direction Selective}
\newacronym[description={DVS optical flow method that fits a plane to the local event cloud}]{lp}{LP}{Local Plane}
\newacronym[description={False Negative Rate; signal that is incorrectly classified as noise}]{fnr}{FNR}{False Negative Rate}
\newacronym[description={False Positive Rate; noise that is incorrectly classified as signal}]{fpr}{FPR}{False Positive Rate}
\newacronym[description={Inter Spike Interval (nomenclature from neuroscience)}]{isi}{ISI}{Inter Spike Interval}
\newacronym[description={Multipurpose block random access memory module in FPGA}]{bram}{BRAM}{Block RAM}
\newacronym[description={Register Transfer Logic intermediate form, consisting of combinational and synchronous register logic cells}]{rtl}{RTL}{Register Transfer Logic}
\newacronym[description={Search Area for block matching}]{sa}{SA}{Search Area}
\newacronym[description={Single Threshold Metric; a measure of the ROC TPR/FPR tradeoff at one discrimination threshold}]{stm}{STM}{Single Threshold Metric}
\newacronym[description={Slice-based FAST that uses accumulated event count slices for detecting keypoints}]{sfast}{SFAST}{Slice-based FAST}
\newacronym[description={Surface of Active Event; image of latest event timestamps, same as Timestamp Image}]{sae}{SAE}{Surface of Active Events}
\newacronym[description={System on Chip; FPGA with embedded programmable processor}]{soc}{SoC}{System on Chip}
\newacronym[description={Timestamp Image; image of latest event timstamps, same as Surface of Active Events}]{ti}{TI}{Timestamp Image}
\newacronym[description={True Negative Rate; noise that is correctly classified as noise}]{tnr}{TNR}{True Negative Rate}
\newacronym[description={True Positive Rate; signal that is correctly classified as signal}]{tpr}{TPR}{True Positive Rate}
\newacronym[description={Visual Odometry}]{vod}{VOD}{Visual Odometry}
\newacronym[longplural={First In First Out memories}]{fifo}{FIFO}{First In First Out memory}
\newacronym{aae}{AAE}{Average Angular Error}
\newacronym{abmof}{ABMOF}{Adaptive Block Matching Optical Flow}
\newacronym{aee}{AEE}{Average Endpoint Error}
\newacronym{aer}{AER}{Address Event Protocol}
\newacronym{aop}{AoP}{Angle of Polarization}
\newacronym{aps}{APS}{Active Pixel Sensor}
\newacronym{asic}{ASIC}{Application Specific Integrated Circuit}
\newacronym{auc}{AUC}{Area Under the Curve}
\newacronym{baf}{BAF}{Background Activity Filter}
\newacronym{ba}{BA}{Background Activity}
\newacronym{bmof}{BMOF}{Block Matching Optical Flow}
\newacronym{bm}{BM}{Block Matching}
\newacronym{cfa}{CFA}{Color Filter Array}
\newacronym{cf}{CF}{Complementary Filter}
\newacronym{cis}{CIS}{CMOS Image Sensor}
\newacronym{cnn}{CNN}{Convolutional Neural Network}
\newacronym{cots}{COTS}{Commodity Off-The-Shelf}
\newacronym{cpu}{CPU}{Central Processing Unit}
\newacronym{cv}{CV}{Computer Vision}
\newacronym{davis}{DAVIS}{Dynamic and Active pixel VIsion Sensor}
\newacronym{dnn}{DNN}{Deep Neural Network}
\newacronym{dof}{DOF}{Degree of Freedom}
\newacronym{dop}{DoLP}{Degree of Linear Polarization}
\newacronym{dr}{DR}{Dynamic Range}
\newacronym{dram}{DRAM}{Dynamic RAM}
\newacronym{drcn}{DRCN}{Deep Recurrent Convolutional Network}
\newacronym{dsp}{DSP}{Digital Signal Processing unit}
\newacronym{dvs}{DVS}{Dynamic Vision Sensor}
\newacronym{dwf}{DWF}{Double Window Filter}
\newacronym{edp}{EDP}{Event Denoising Precision}
\newacronym[description={Extinction Ratio (the reciprocal of the ratio of cross-polarized light that passes the linear polarizer}]{er}{ER}{Extinction Ratio}
\newacronym{fom}{FOM}{Figure of Merit}
\newacronym{fpga}{FPGA}{Field Programmable Gate Array}
\newacronym{fpn}{FPN}{Fixed Pattern Noise}
\newacronym{fps}{FPS}{Frames Per Second}
\newacronym{fsae}{FSAE}{Filtered Surface of Active Events}
\newacronym{fwf}{FWF}{Fixed Window Filter}
\newacronym{gpu}{GPU}{Graphics Processing Unit}
\newacronym{gt}{GT}{Ground Truth}
\newacronym{hdl}{HDL}{Hardware Description Language}
\newacronym{hdr}{HDR}{high dynamic range}
\newacronym{hls}{HLS}{High Level Synthesis}
\newacronym{icm}{ICM}{Iterated Conditional Modes}
\newacronym{id}{ID}{Index Decay}
\newacronym{iir}{IIR}{Infinite Impulse Response}
\newacronym{imu}{IMU}{Inertial Measurement Unit}
\newacronym{inceptiveevent}{IE}{Inceptive Event}
\newacronym{iot}{IoT}{Internet of Things}
\newacronym{ip}{IP}{Intellectual Property}
\newacronym{its}{ITS}{Invariant Time Surface}
\newacronym{li}{LI}{Leaky Integrator}
\newacronym{lk}{LK}{Lucas-Kanade}
\newacronym{mpeg}{MPEG}{Motion Picture Experts Group}
\newacronym{na}{NA}{Numerical Aperture}
\newacronym{nnb}{NNb}{Nearest Neighbor}
\newacronym{of}{OF}{Optical Flow}
\newacronym{onf}{ONF}{Order(N) Filter}
\newacronym{pcb}{PCB}{Printed Circuit Board}
\newacronym{pd}{PD}{photodiode}
\newacronym{pdavis}{PDAVIS}{Polarization Dynamic and Active pixel VIsion Sensor}
\newacronym{pe}{PE}{Processing Element}
\newacronym{pfa}{PFA}{Polarization Filter Array}
\newacronym{pl}{PL}{programmable Logic}
\newacronym{por}{POR}{Positive Output Ratio}
\newacronym{prm}{PRM}{Pixel Rendering Module}
\newacronym{ps}{PS}{Processing System}
\newacronym{pugm}{PUGM}{Probabilistic Undirected Graph Model}
\newacronym{qwp}{QWP}{Quarter Wave Plate}
\newacronym{ram}{RAM}{Random Access Memory}
\newacronym{ratp}{RATP}{Recursive Adaptive Temporal Pooling}
\newacronym{roc}{ROC}{Receiver Operating Characteristic}
\newacronym{roi}{ROI}{Region of Interest}
\newacronym{rpmd}{RPMD}{Relative Plausibility Measure of Denoising}
\newacronym{rpm}{RPM}{Revolutions per Minute}
\newacronym{sad}{SAD}{Sum of Absolute Differences}
\newacronym{sd}{SD}{Secure Digital}
\newacronym{silc}{SILC}{Speed Invariant Learned Corners}
\newacronym{sits}{SITS}{Speed Invariant Time Surface}
\newacronym{slam}{SLAM}{Simultaneous Localization And Mapping}
\newacronym{sm}{SM}{Supplementary Material}
\newacronym{snr}{SNR}{Signal to Noise Ratio}
\newacronym{sram}{SRAM}{Static RAM}
\newacronym{stcf}{STCF}{SpatioTemporal Correlation Filter}
\newacronym{susan}{SUSAN}{Smallest Univalue Segment Assimilating Nucleus)}
\newacronym{tda}{TDA}{Time Decay Adapted}
\newacronym{td}{TD}{Time Decay}
\newacronym{timsl}{TS}{time slice}
\newacronym{usb}{USB}{Universal Serial Bus}
\newacronym{vga}{VGA}{Video Graphics Adaptor}
\newacronym{vhdl}{VHDL}{Very High-Speed Integrated Circuit Hardware Description Language}
\newacronym{zoh}{ZOH}{Zero-Order Hold}

\newcommand{\fcorner}{{f_\text{3dB}}}

\usepackage{xr-hyper} % needed for xref'ing to SM

\newcommand*{\myexternaldocument}[1]{%
    \externaldocument{#1}%
    \addFileDependency{#1.tex}%
    \addFileDependency{#1.aux}%
}
\begin{document}

\myexternaldocument{SupplementalMaterials}

\title{Bio-inspired Polarization Event Camera}
\author[1,2,$\dagger$]{Germain Haessig}
\author[2,3,$\dagger$]{Damien Joubert}
\author[4,$\dagger$]{Justin Haque}
\author[4]{Yingkai Chen}
\author[2,3]{Moritz B. Milde}
\author[2,*]{Tobi Delbruck}
\author[4,*]{Viktor Gruev}
\affil[1]{\footnotesize{AIT Austrian Institute of Technology, Center for Vision, Automation \& Control, High-performance Vision Systems, Vienna, Austria}}
\affil[2]{\footnotesize{Institute of Neuroinformatics, University of Zurich and ETH Zurich, Switzerland}}
\affil[3]{\footnotesize{International Centre for Neuromorphic Systems, The MARCS Institute, Western Sydney University, Sydney, Australia}}
\affil[4]{\footnotesize{Department of Electrical and Computer Engineering, University of Illinois at Urbana-Champaign, Urbana, IL, USA}}
\affil[$\dagger$]{\footnotesize{These authors contributed equally}}
\affil[*]{\footnotesize{Contact authors emails: tobi@ini.uzh.ch, vgruev@illinois.edu}}

%TC:ignore
\maketitle
\newrefsection
%TC:endignore

\glsunset{pdavis} % use up definition for body text.   It is defined in the SM
\glsunset{rpm} % use up definition for body text.   It is defined in the SM
\section{Abstract}
    The stomatopod (mantis shrimp) visual system ~\autocite{Marshall88,cronin2014visual,Horvath2014-polarization-light-and-vision-in-animals} has recently provided a blueprint for the design of paradigm-shifting polarization and multispectral imaging sensors ~\autocite{Altaquieabe3196,GarciaPolarization,BioinspiredSenorsReview2021,jen2011biologically,LiuEndoscope,Blaireaaw7067, ZhangPolarization}, enabling solutions to challenging medical~\autocite{LiuEndoscope,Blaireaaw7067} and remote sensing problems~\autocite{Powelleaao6841}. However, these bioinspired sensors lack the \tip{hdr} and asynchronous polarization vision capabilities of the stomatopod visual system, limiting temporal resolution to $\sim$12\,ms and dynamic range to $\sim$72\,dB. Here we present a  novel stomatopod-inspired polarization camera which
    mimics the sustained and transient biological visual pathways to save power and sample data beyond the maximum Nyquist frame rate.
    This bio-inspired sensor simultaneously captures both synchronous intensity frames
    and asynchronous  polarization brightness change information with sub-millisecond latencies over a million-fold range of illumination. Our \tip{pdavis} camera is comprised of 346x260 pixels, organized in 2-by-2 macropixels, which filter the incoming light with four linear polarization filters offset by 45\textdegree. Polarization information is reconstructed using both low cost and latency event-based algorithms and more accurate but slower deep neural networks. 
    Our sensor is used to image \tip{hdr} polarization scenes which vary at high speeds and to observe dynamical properties of single collagen fibers in bovine tendon under rapid cyclical loads. 

\section{Main}

\begin{figure}[!htp]
\centering
\includegraphics[width=\textwidth]{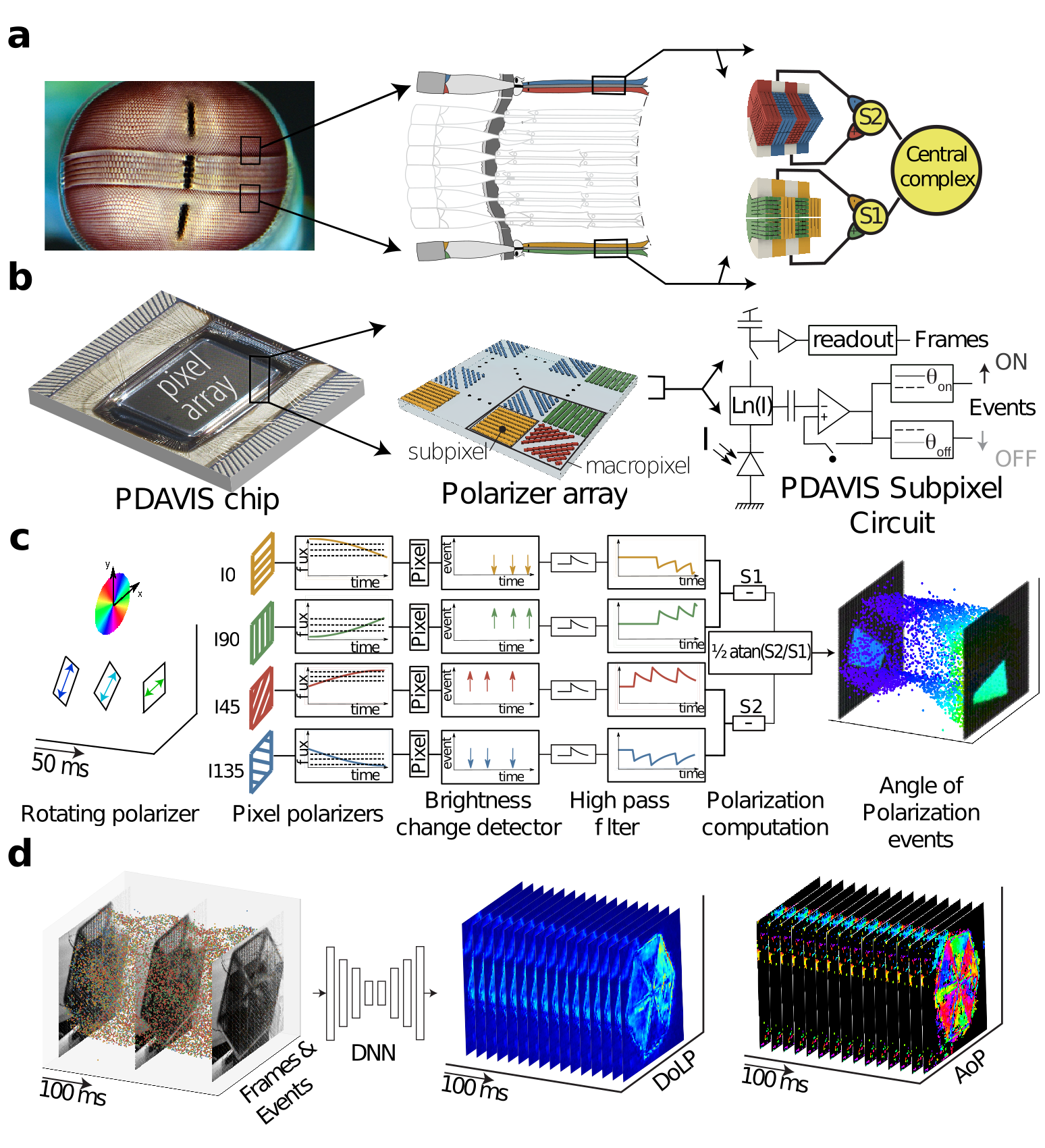}
\captionsetup{width=\textwidth}
\caption{
\textbf{Overview of our bio-inspired polarization vision sensor. }{\normalsize a,} Polarization vision in the mantis shrimp eye (left) is in part enabled by two sets of orthogonal microvilli located in the dorsal and ventral hemisphere (center), capturing total of 4 linear polarization states offset by 45 degrees (right). This polarization sensitivity paired with logarithmic photoreceptors that output only brightness change enable the mantis shrimp to be effective predator in the shallow coral reefs~\autocite{cronin2014visual}. 
{\normalsize b,} The \tipshort{pdavis} polarization event camera (left), effectively mimics the mantis shrimp eye by integrating an array of pixelated polarization filters offset by 45 degrees (center, angle indicated by false colors) with a vision sensor that provides both sustained pathway frames and transient pathway log-scale brightness change events (right) (see \ref{sup:pdavis_camera}).
{\normalsize c,} A rectangular rotating linear polarizer (left) generates a stream of brightness change events from the four subpixels in the \tipshort{pdavis} macropixels that see the polarizer (center). A temporal filter approximates the temporal derivative from the four individual pixels and then computes the \tipshort{aop}. The result is a stream of \tipshort{aop} events with low latency (right). (see \ref{sup:reconstruction_from_pdavis}).
{\normalsize d,}~A polarization filter wheel is rotated in front of \tip{pdavis}, which produces frames and events (left) represented in false colors shown in (c). A \tipshort{dnn} (center-left) reconstructs \tipshort{dop} (center-right) and \tipshort{aop} (right) from the brightness change events at a higher rate than the cameras maximum frame rate (see \ref{sup:reconstruction_from_pdavis}).
} 
\label{fig:fig1_header}
\end{figure}

Visual information is encoded in light by intensity, color, and polarization~\autocite{cronin2014visual}. This information is sensed by biological eyes and artificial cameras which each have been optimized by evolution driven by maximum fitness. Eyes have evolved to support visually guided behavior for the benefit of survival, while digital cameras have mainly evolved to supply consumer demand for high-resolution photography. These different evolutionary paths have created very different visual systems. Existing spectral and polarization digital cameras use synchronous and generally redundant frames with linear photo response~\autocite{Blaireaaw7067,GarciaPolarization, MukulPolarization, TokudaPolarization, HsuPolarization}. 
By  contrast, eyes are asynchronous, have a compressed nonlinear response, and their output is sparse and highly informative~\autocite{cronin2014visual}. 

The mantis shrimp visual system (Fig.~\ref{fig:fig1_header}a) is considered one of the most sophisticated visual systems in nature. It is sensitive to more than 12 spectral, 4 linear, and 2 circular polarization channels~\autocite{Marshall88, cronin2014visual}. Its photosensitive microvilli have a logarithmic \tip{hdr} response to incident light.  
Sensitivity to linearly polarized light is in part expressed in the dorsal and ventral parts of the ommatidia, where individual photoreceptors are comprised of orthogonal sets of microvilli sensitive to orthogonal polarization states. The dorsal/ventral views largely overlap, and since the dorsal and ventral microvilli are offset by 45\textdegree, four linear polarization states offset by 45\textdegree\ are captured by the eye. The logarithmic photo responses of the microvilli enable high dynamic range polarization sensing capabilities, while their asynchronous response to temporally varying brightness greatly reduces the visual information that is transmitted to their brain for further processing.
It is believed that  mantis shrimp use polarization to discriminate short-range prey~\autocite{cronin2014visual}, to select a mating partner~\autocite{marshall2019polarisation} and to orient during short-range navigation using celestial polarization patterns~\autocite{PATEL20201981}.

Our work capitalizes on the development of bioinspired neuromorphic vision sensors, which have enabled higher dynamic range and lower latency machine vision~\autocite{brandli14,Gallego2020-survey-paper}.
Inspired by the ommatidia of mantis shrimp, individual \tip{pdavis} subpixel circuits~\autocite{brandli14} are each overlaid with one of four pixelated linear polarization filters (Fig.~\ref{fig:fig1_header}b). 
The \tip{pdavis} takes inspiration from biology by saving energy by partitioning the perception of fine detail and fast motion into sustained and transient pathways~\autocite{cronin2014visual}.
It provides a relatively low frequency synchronous readout of frames like conventional cameras (the ``sustained'' pathway), 
and it concurrently outputs a high frequency stream of asynchronous brightness change events (the ``transient'' pathway).
Each event represents a signed log intensity change. 
Pixels that see more brightness change generate more events, and the events have sub-millisecond temporal resolution driven by the dynamics of the scene. The events enable reconstructing the absolute intensity between the synchronous frame intensity samples.

\section{Reconstructing Polarization Information}

Polarization is encoded in the relative responses of the subpixels with the 4 linear polarizers offset by 45\textdegree. Absolute intensity is encoded in the sum of crossed polarization subpixel outputs. For a direct comparison with a state-of-the-art commercially available polarization imaging sensor (Sony IMX250~\autocite{Sony2021-triton-imx250}), we processed the \tip{pdavis} output to extract \tip{aop}, \ie the predominant axis of oscillation as light propagates in time and space, and \tip{dop}, \ie the amount of linearly polarized in the incident light. We studied three related algorithms: an economical algorithm processing only events, an economical \tip{cf} algorithm which fuses frames and events~\autocite{Scheerlinck2019-complementary-filter}, and an expensive \tip{dnn} that takes events as input~\autocite{scheerlinck2020fast} (see Methods, \ref{sup:reconstruction_from_pdavis} and Supplementary Table \ref{tab:comparing_reconstruction}).
We did four sets of experiments. 

\begin{figure}[!htp]
\centering
\includegraphics[width=1\columnwidth]{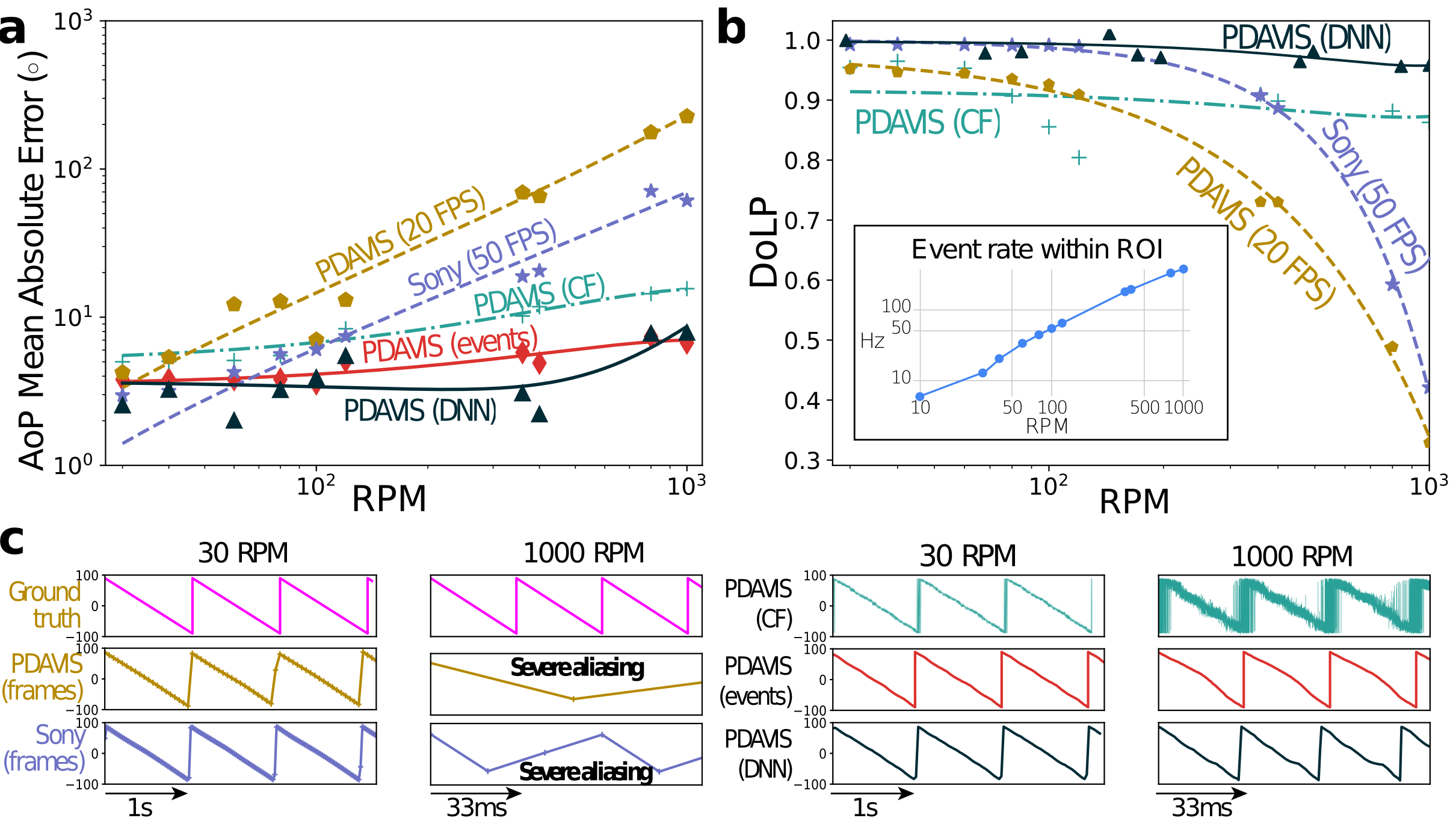}
\caption{\textbf{Comparison of frame-based and event-based \tip{aop} and \tip{dop} reconstruction accuracy at various rotational speeds.} 
Input is image of fully linearly polarized light from a rotating linear polarizer.
\textbf{a}~Mean absolute error (\textbf{MAE}) of \tip{aop} reconstruction for the various methods averaged over 12x12 \tip{roi} centered on the polarizer. (see \ref{sup:reconstruction_from_pdavis}). 
\tip{pdavis} and Sony frame rates are shown in plots and both exposure durations were fixed to 20\,ms.
The \textit{frames} methods use only synchronous intensity frames. 
The \tip{cf} method fuses 20\,FPS frames and events using a method adapted from \autocite{Scheerlinck2019-complementary-filter}. The \textit{events} method is illustrated in Fig.~\ref{fig:fig1_header}c. 
The \tip{dnn} method uses only events together with the convolutional recurrent neural network \autocite{scheerlinck2020fast}.
\textbf{b}~\tip{dop} reconstructions. Both \tip{cf} or \tip{dnn} methods show that using events allows reconstructing \tip{dop} well beyond the limiting Nyquist frequency of the frame sampling. 
Inset plots the event rate per pixel within the polarizer \tipshort{roi} versus \tipshort{rpm}; the event rate outside \tipshort{roi} is $<1$\,Hz after denoising.
\textbf{c}~Reconstruction of the \tip{aop} from a 100 pixel \tipshort{roi} using various methods. 
The frame based Sony and \tip{pdavis} reconstructions are severely aliased at 1000\,RPM. which is not true for any of the \tip{pdavis} methods using events.
} 
\label{fig:fig2_aop_reconstruction}
\end{figure}

First, we assessed the ability to reconstruct the time-varying \tip{aop} of fully linearly polarized light (with \tip{dop}=1) by rotating a linear polarizer at constant speeds (Fig.~\ref{fig:fig2_aop_reconstruction}, \ref{sup:pdavis_camera}, Supplementary Figs.~\ref{fig:characterization_bench}, Supplementary Video 1). At low rotation speeds of less than 60\,\tip{rpm}, the \tip{aop} reconstruction error from both sensors is less than 5\textdegree, with the Sony sensor having the lowest reconstruction error. The reconstructed \tip{dop} is nearly 1 from both sensors, as expected. 
Since Sony’s polarization sensor is fabricated in an optimized semiconductor fab, the mismatches in both the optical properties of the pixelated polarization filters and electrical properties of the photodiodes and read-out circuits are minimal, resulting in high accuracy in the reconstructed polarization information for slow rotation rate. Our \tip{pdavis} prototype has larger mismatches between the optical properties of the pixelated filters as well as read-out electronics resulting in larger error at low frequency, which can be mitigated by calibration~\autocite{gruev10}. However, when the linear polarizer is rotated above $\sim$100 RPM, the Sony and \tip{pdavis} frames start aliasing and motion blurring (Fig.~\ref{fig:fig2_aop_reconstruction}c), which decreases the estimated \tip{dop} and increases the \tip{aop} error.
The \tip{pdavis} events maintain precise timing information and the reconstructions using events have \tip{aop} and \tip{dop} error less than 10\textdegree\ all the way to 1000\,\tipshort{rpm}.

\begin{figure}[!htp]
\centering
\includegraphics[width=1\columnwidth]{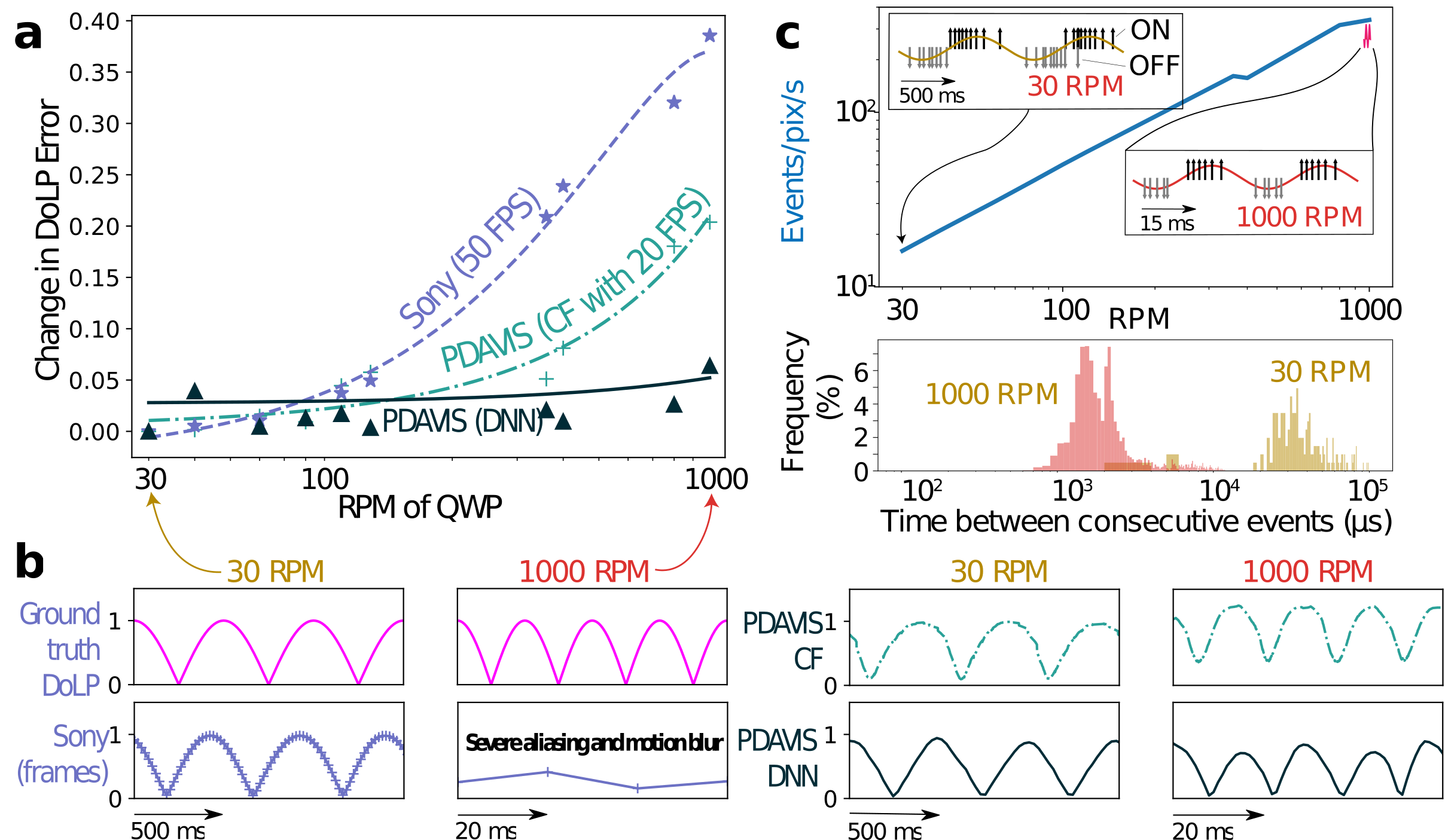}
\caption{\textbf{Comparison of frame-based and event-based reconstruction of \tip{dop} with various rotation speeds of a \tip{qwp}.}  Each cycle of \tip{qwp} rotation produces four cycles of \tip{dop}.
\textbf{a}~Growth of the mean absolute error of \tip{dop} reconstruction with RPM. The Sony has only synchronous intensity frames (see \ref{sup:conv_pola_imaging}). The \tip{cf} method fuses \tip{pdavis} frames and events (see \ref{sup:cf}). 
The \tip{dnn} method uses only events (see \ref{sup:firenet}).
\textbf{b}~Reconstruction of the absolute \tip{dop} from a 100 pixel \tip{roi} using various methods. The frame based Sony reconstruction is most accurate at low frequency but is severely aliased at 1000\,RPM, which is not true for any of the \tip{pdavis} methods using events.
\textbf{c} Statistics of events. Upper plot is event rate versus \tip{qwp} \tipshort{rpm}. Insets show actual ON and OFF events from a subpixel in response to the sinusoidal intensity variation. Lower plot shows histograms of the time between two consecutive events for the two rotational speeds of the \tip{qwp}.} 
\label{fig:fig3_dop_reconstruction}
\end{figure}

% In the first experiment, we used fully polarized light. 
The second experiment (Fig.~\ref{fig:fig3_dop_reconstruction}, Supplementary Video 2) assessed the ability to measure time-varying \tip{dop} while \tip{dop} and \tip{aop} both vary with time. We combined a rotating linear polarizer with a fixed \tip{qwp}.
Fig.~\ref{fig:fig3_dop_reconstruction}a shows how much the \tip{dop} error increases as a function of the speed of the \tip{qwp}, 
compared to 30 \tip{rpm}. 
For visual comparison, each method is defined to have zero ``change of error'' at the lowest frequency.
Fig.~\ref{fig:fig3_dop_reconstruction}b shows the absolute \tip{dop} measured by each method for 30 and 1000\,\tipshort{rpm}.
At low RPM, the Sony camera makes the most accurate estimate of \tip{dop}.
When the \tip{qwp} is rotated at higher speeds, 
the frames from both cameras become aliased and motion blurred, 
resulting in a large error increase of over 50\% in the Sony \tip{dop}; at 1000\,RPM, the Sony frames are hopelessly blurred and aliased (Fig.~\ref{fig:fig3_dop_reconstruction}b, Sony (frames)). 
However, the \tip{cf} method fuses the \tip{pdavis} events with its 20\,FPS frames, clearly improving the reconstruction in comparison with the 50\,FPS Sony (Fig.~\ref{fig:fig3_dop_reconstruction}b, \tip{pdavis} CF). 
Finally, using only the \tip{pdavis} events with the \tip{dnn} method keeps the growth in reconstruction error below 8\% all the way to 1000\,RPM
 (Fig.~\ref{fig:fig3_dop_reconstruction}b, \tip{pdavis} DNN).
Fig.~\ref{fig:fig3_dop_reconstruction}c shows the statistics of the events. 
At 30\,RPM, the distribution of interevent time intervals (lower histogram) shows that the events are widely spaced because the brightness changes are slow.
At 1000\,RPM, the distribution moves to much shorter event intervals, down to less than 1\,ms. 
The event rate (Fig.~\ref{fig:fig3_dop_reconstruction}c upper plot) is directly proportional to RPM. 
The insets of the event rate plot show events from one pixel; 
the structure of ON and OFF events is similar for 30\,RPM and 1000\,RPM, but speeds up by a factor of 30. 
This low latency asynchronous \tip{pdavis} output allows the measurement of fast brightness changes, 
which occur much more rapidly than the fixed frame rate; 
the \tip{pdavis} events sample as needed, up to more than 1\,kHz in this experiment.

\begin{figure}[!htp]
\centering
\includegraphics[width=1\columnwidth]{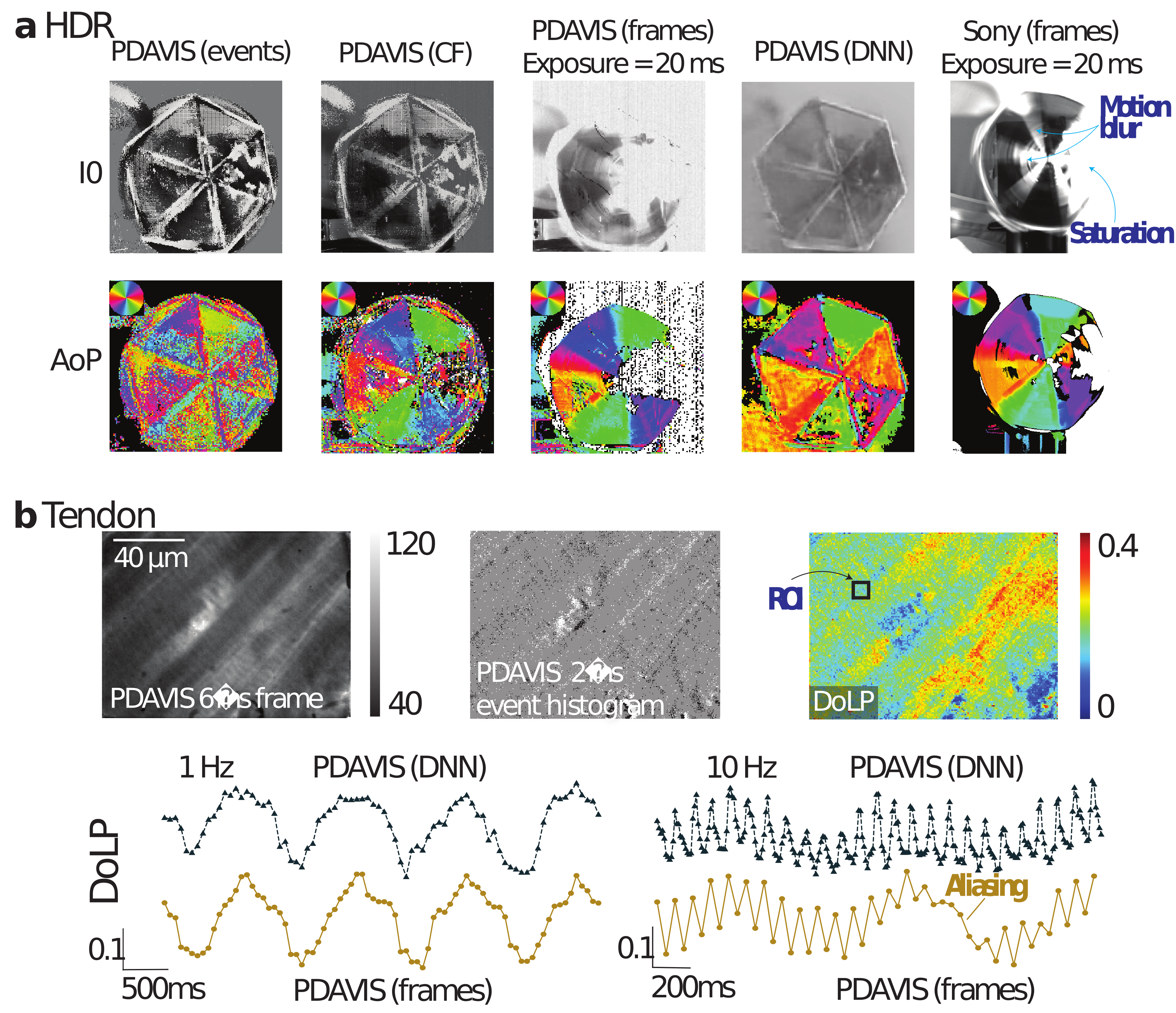}
\caption{\textbf{Applications of \tip{pdavis} for remote sensing and scientific imaging.} 
\textbf{a}~Images of a rotating fan polarizer under \tip{hdr} illumination captured by the \tip{pdavis} and Sony's camera using various reconstruction methods (see \ref{sup:reconstruction_from_pdavis}). 
The fan is constructed from triangular linear polarizers arranged with 30\textdegree\ \tip{aop} spacing.
The illumination ratio between bright and dark parts is 2000 and the rotation speed is 200\,RPM.
The \textsf{I$_0$} images show reconstructed monochromatic intensity from pixels with 0 degree pixelated polarization filters of Fig.~\ref{fig:fig1_header}.
The weakly-polarized regions in the \tip{aop} images are masked out using the measured \tip{dop}.
\textbf{b}~Imaging a tendon that is periodically stretched at 1\,Hz and 10\,Hz. The images (upper half) show a single frame (left), an event time window (center) and the corresponding \tip{dop} reconstruction result (right) using the \tip{dnn} method (see \ref{sup:bovine} and \ref{sup:reconstruction_from_pdavis}).
The event time window is rendered from a fixed time interval where ON (white) and OFF (black) events are accumulated to the starting gray image.
The \tip{dnn} input is 3D event tensors (the frames are not used) that have a duration of 50 ms or 10 ms, resulting in output at 20 or 100 \tipshort{fps} for 1\,Hz and 10\,Hz respectively.
The traces (lower half) compare the \tip{dnn} and frame-based reconstructed \tip{dop}, averaged over the \tip{roi} indicated in the \tip{dop} image (upper right). With 1\,Hz stretching (left), both methods yield similar results. With 10\,Hz stretching (right), the \tip{dnn} reconstructs the \tip{dop} while the 20\,\tipshort{fps} frame-based reconstruction is severely aliased.
}
\label{fig:fig4_fast_hdr_and_tendon}
\end{figure}

The third experiment (Fig.~\ref{fig:fig4_fast_hdr_and_tendon}a, Supplementary Videos 3 and 4) compares the \tip{pdavis} and Sony dynamic range. 
We imaged set of polarization filters offset by 30\textdegree rotating at 200\,\tipshort{rpm} (5\,rev/s) under high contrast 2000:1 lighting, such as commonly encountered in remote sensing of natural environments. 
The Sony camera exposure is set to 20\,ms to capture the darker part of the scene without underexposing it, which overexposes and saturates the brighter part, preventing \tip{aop} measurement. The large motion blur is  visible in the I0 image and incorrect \tip{aop} in the blurred regions.
The \tip{pdavis} can measure the \tip{aop} in both lighting conditions. Even though the \tip{pdavis} frame is also motion blurred, all event-based methods produce sharp images.
% Even with this relatively densely moving input scene, the \tip{pdavis} event data rate is 7X smaller than it would be from an equivalent resolution 1000\,\tipshort{fps} image sensor.

The fourth experiment (Fig.~\ref{fig:fig4_fast_hdr_and_tendon}b, Supplementary Video 5) shows a potential medical imaging application of the \tip{pdavis} (see Methods and \ref{sup:bovine}). We imaged the dynamics of a bovine tendon subjected to cyclical stress, such that its birefringent properties are time varying. As circularly polarized light transmits through the tendon, it becomes elliptically polarized and the \tip{dop} provides an indirect measurement of the birefringent properties, assuming minimal scattering. Using high optical magnification, we can observe strain patterns over time of the individual collagen fibers that comprise the tendon. Due to the high optical magnification and rapid movement of the collagen fibers, preventing aliasing would require a frame rate that is a large multiple of the cycle rate. By contrast, \tip{dnn} reconstruction of the \tip{dop} (using only events) provides measurement of the dynamic properties of the individual collagen fibers at higher frequency than the maximum frame rate.

\section{Discussion}

Airborne, underwater, and space-based applications can require high temporal resolution and \tip{hdr}, together with spectral and polarization sensitivity. All of these requirements increase the data rate, but the bioinspired sparse data streams and local gain control of event cameras enables near-sensor processing with low latency and a small computational footprint together with \tip{hdr}.
To compare \tip{pdavis} with state-of-the-art polarization sensors, we developed novel event-driven reconstruction algorithms and compared their angle and degree of polarization reconstruction abilities to the frame-based camera reconstruction.
The pure event-driven algorithm is the most economical, but it cannot reconstruct the degree of linear polarization, which requires an estimate of the absolute intensity which the event stream does not provide.
The \tip{cf} helps to overcome this limitation by fusing the event stream with periodically captured frames while only slightly increasing the computational cost.
The \tip{dnn} provides the most accurate reconstruction, but requires  a power hungry and expensive \tipshort{gpu} for real time operation, which may not be affordable in a remote environment close to the sensor or with minimum latency.
A limitation of the \tip{pdavis} is with dense scenes, which can saturate the event output capacity, causing event loss. In these situations, a conventional frame-based polarization camera could be better suited. 
By adopting a bioinspired combination of sustained and transient pathway, the \tip{pdavis} bridges a gap between the limited temporal and dynamic range of conventional frame-based polarization cameras and complex solid state imagers\autocite{Etoh_2011-ultra-hs-ccd} or streak cameras\autocite{Gao2014-cup-10to11fps-streak-cam} that can record short sequences at $>10^7$ \tipshort{fps}. 
This gap is normally filled by high frame rate cameras that consume a lot of power and demand bright lighting for the short exposure times.
The \tip{pdavis} enables continuous \tip{aop} and \tip{dop} measurement with high contrast illumination at frequencies several times the Nyquist rate of frame-based image sensors. 
The \tip{pdavis} event output triggers data acquisition and processing only when needed making it ideally matched with the increasing development of activation-sparsity aware neural accelerators\autocite{Davies2018-loihi,Chen2017-eyeriss,Aimar2018-nullhop}.

\section{Methods}

The \tip{pdavis} \tip{pfa} is fabricated on a quartz wafer using interference lithography and reactive ion etching. Reactive ion etching transfers the photoresist pattern to the silicon dioxide, which serves as a hard mask for etching the underlying aluminum. The filter array is then laser cut to match the total size of the pixel array. The individual polarization filters are comprised of aluminum nanowires with 70\,nm width, 70\,nm spacing, and 150\,nm height, and have extinction ratio of ~40 at 500\,nm incident light (see Methods, \ref{sup:pdavis_camera}, and Supplementary Figs.~\ref{fig:pfa_assembly} and ~\ref{fig:er_plot}). The \tip{pfa} and pixel array are aligned using a 6-\tipshort{dof} positioning stage and are bonded using ultraviolet-curing optical adhesive. (see Supplementary Table \ref{tab:spec_table}). 

% The polarization filter array is fabricated by first depositing 150\,nm thick aluminum on a quartz glass followed by 30\,nm silicon dioxide. Interference UV lithography selectively patterns a photoresist mask with 70-nm line width and 70-nm spacing at four different orientations offset by 45 degrees.  The polarization filters are aligned and attached to the packaged \tipshort{davis} sensor using a custom-built 6-\tipshort{dof} alignment stage combined with collimated linearly polarized light (see \ref{sup:pfa_assembly} and Figs.~\ref{fig:pfa_assembly},~\ref{fig:characterization_bench}).

A custom PCB hosts the \tip{pdavis}  chip and streams data (frames and events) to the host computer (see \ref{sup:davis} and Supplementary Fig. \ref{fig:davis_pixel}). 
To recover and reconstruct information about \tip{aop} and \tip{dop} we compared 4 different \tip{pdavis} methods (for an overview see Supplementary Table~\ref{tab:comparing_reconstruction}, \ref{sup:reconstruction_from_pdavis}). It also enabled a comparison with Sony's camera (see Supplementary Table~\ref{tab:spec_table}, Figs.~\ref{fig:er_plot},\ref{fig:dr_plot}).
We applied conventional frame-based reconstruction, purely event-driven reconstruction, event-frame sensor fusion, and a deep recurrent convolutional neural network (see \ref{sup:reconstruction_from_pdavis}, Fig.~\ref{fig:cf_transfer_function}).
For Fig.~\ref{fig:fig4_fast_hdr_and_tendon}b, we sectioned bovine tendons and mounted them on a custom-built stage to enable cyclical loading, and imaged them with a microscope objective (see \ref{sup:bovine}).
% x10 optical lens with a numerical aperture of 0.25 .

\section{Data availability}
The data that support the findings of this study are available within the article and its Supplementary Information. Raw data collected from the sensors are available upon request from V.G. or T.D.

\section{Code availability}
Software code and data used to generate polarization data are available online. The link will be made available once the paper is accepted for publication. 

%TC:ignore
\section{Funding}
Fabrication of the polarization filters and travel was partially funded by the Swiss National Science Foundation Sinergia projects \#CRSII5-18O316, \#CRSII5-18O316 and ONR Global-X N62909-20-1-2078. This material is based upon research supported by, or in part by, the U. S. Office of Naval Research (N62909-20-1-2078, N00014-19-1-2400 and N00014-21-1-2177) and U.S. Air Force Office of Scientific Research (FA9550-18-1-0278).

\section{Acknowledgments}

The authors thank Greg Cohen, Tom Cronin, and Andrey Kaneev for comments; Cedric Scheerlinck for help with implementing the \tip{cf}; Steven Blair and Alex Pietros for help with aligning and bonding the polarization filter arrays to the DAVIS sensors; Steven Blair, Zuodong Liang, Colin Symons, and Zhongmin Zhu for help with measurements; Steven Blair and Zhongmin Zhu for help with data analysis; and Leanne Iannucci and Spencer Lake for assistance with the tendon experiments. 

\section{Author Contributions}
T.D. and V.G. conceived the idea and supervised the project. G.H. and V.G. aligned and bonded the polarization filter arrays to the DAVIS sensors. 
DAVIS sensor was developed by T.D lab. 
Measurements were performed by D.J., Y.C., and J.H.
Complementary filter was developed by T.D., G.H. and D.J. DNN data analysis was performed by J.H and Y.C. G.H., D.J., T.D., and M.B.M developed the event-based reconstruction algorithm. All authors contributed to the data analysis and writing of the paper.

%\bibliographystyle{unsrt}
%\bibliography{pDAVIS}

\printbibliography
% \pagebreak
% \appendix{Supplemental Materials}
% \begin{center}
% \textbf{\large SUPPLEMENTARY INFORMATION}
% \end{center}
%%%%%%%%%% Merge with supplemental materials %%%%%%%%%%
%%%%%%%%%% Prefix a "S" to all equations, figures, tables and reset the counter %%%%%%%%%%
\setcounter{section}{0}
\setcounter{equation}{0}
\setcounter{figure}{0}
\setcounter{table}{0}
\setcounter{page}{1}
\setcounter{footnote}{0}
\setcounter{enumiv}{1}
\makeatletter
\renewcommand{\thesection}{Supplementary Material~\arabic{section}}
\renewcommand{\theequation}{S\arabic{equation}}
\renewcommand{\thefigure}{S\arabic{figure}}
%TC:ignore
\renewcommand{\thefootnote}{\textit{\alph{footnote}}}  % https://tex.stackexchange.com/questions/127270/footnotes-with-italicized-letters
\glsresetall % redfine for all acyronyms for SM

\lhead{PDAVIS: Supplementary Materials} 

\section*{Supplementary Materials}

\newrefsection

{
\begin{itemize}
\setlength{\itemsep}{0pt}
    \item\textbf{ \ref{sup:pdavis_camera}} provides details of the \tip{pdavis} pixel circuit, characterization setup, \tip{pfa} assembly, and specifications of the \tip{pdavis} and Sony cameras.
    \item \textbf{\ref{sup:reconstruction_from_pdavis}} provides details of \tip{pdavis} algorithms for reconstructing polarization.
    \item \textbf{\ref{sup:bovine}} describes the setup of the bovine tendon experiment. 
    \item \textbf{Supplementary Video 1:} The video shows both the frame and event based polarization reconstruction when a linear polarization filter is rotated in front of the PDAVIS camera. The top half of the video represents data acquired from the PDAVIS frames and the bottom half corresponds to event based reconstruction using complementary filter. The left half of the video depicts the intensity data from the four pixelated polarization filters. The right half of the video depicts the reconstructed angle of polarization. It can be observed that the frame based reconstruction (top right) updates \tip{aop} information at much slower speed compared to the event based reconstruction (bottom right). This video demonstrates the aliasing problems associated with frame based polarization imaging.  
     
      \item \textbf{Supplementary Video 2:} The video shows both the frame and event based polarization reconstruction when a \tip{qwp}  filter is rotated in front of the \tip{pdavis} camera. The top half of the video represents data acquired from the \tip{pdavis} frames and the bottom half corresponds to event based reconstruction using the complementary filter. The left half of the video depicts the intensity data from the four pixelated polarization filters. The right half of the video depicts the reconstructed \tip{dop}. It can be observed that the frame based (top right) updates \tip{dop} information at much slower speed compared to the event based reconstruction (bottom right). This video demonstrates the aliasing problems associated with frame based polarization imaging. 
    
        \item \textbf{Supplementary Video 3:} The video shows a high dynamic range scene comprised of six linear polarization filters rotating at \textasciitilde 105\,RPM. The frame based data from both Sony and PDAVIS cannot fatefully reconstruct the \tip{aop} across the entire scene due to the limited dynamic range. The event based reconstruction methods leverage the high dynamic range capabilities of the PDAVIS and reconstruct \tip{aop} information across the entire scene. Furthermore, slight motion blur can be observed in both intensity and \tip{aop} images in the Sony sensor. 
       
          \item \textbf{Supplementary Video 4:} The video shows motion blur problems associated with frame based polarization sensor. The filter wheel is rotated at \textasciitilde 1000\,RPM (16.7\,rev/s), causing severe motion blur in the Sony polarization sensor. Due to the "on-demand" imaging capability provided by events in the PDAVIS, motion blur in this experiment is non existent. 
          \item \textbf{Supplementary Video 5:} Imaging single fibers of a bovine tendon under cyclical load of 1\,Hz and 10\,Hz. The frame based method can accurately monitor the changes of stress in single tendon fibers based on the \tip{aop} information. However, at 10\,Hz cyclical load this information is severely aliased. Event based reconstruction provides updates only at the location of the fibers and provides accurate stress information at the 10\,Hz cyclical load. 
    
\end{itemize}
}
\newpage

\section{PDAVIS camera}
\label{sup:pdavis_camera}

\subsection{PDAVIS chip}
\label{sup:davis}

The \tipshort{davis} chips used to build the \tip{pdavis} cameras were fabricated by Towerjazz Semiconductors in their Fab 2 (Migdal HaEmek, Israel), a 180nm 6-metal \tip{cis} process with optimized buried photodiodes, antireflection coating, and customized microlenses for large pixels. 
The chips are packaged in a ceramic PGA package with a taped glass lid, which we remove for \tip{pfa} assembly (\ref{sup:pfa_assembly}).

\begin{figure*}[!h]
\centering
\includegraphics[width=1.0\columnwidth]{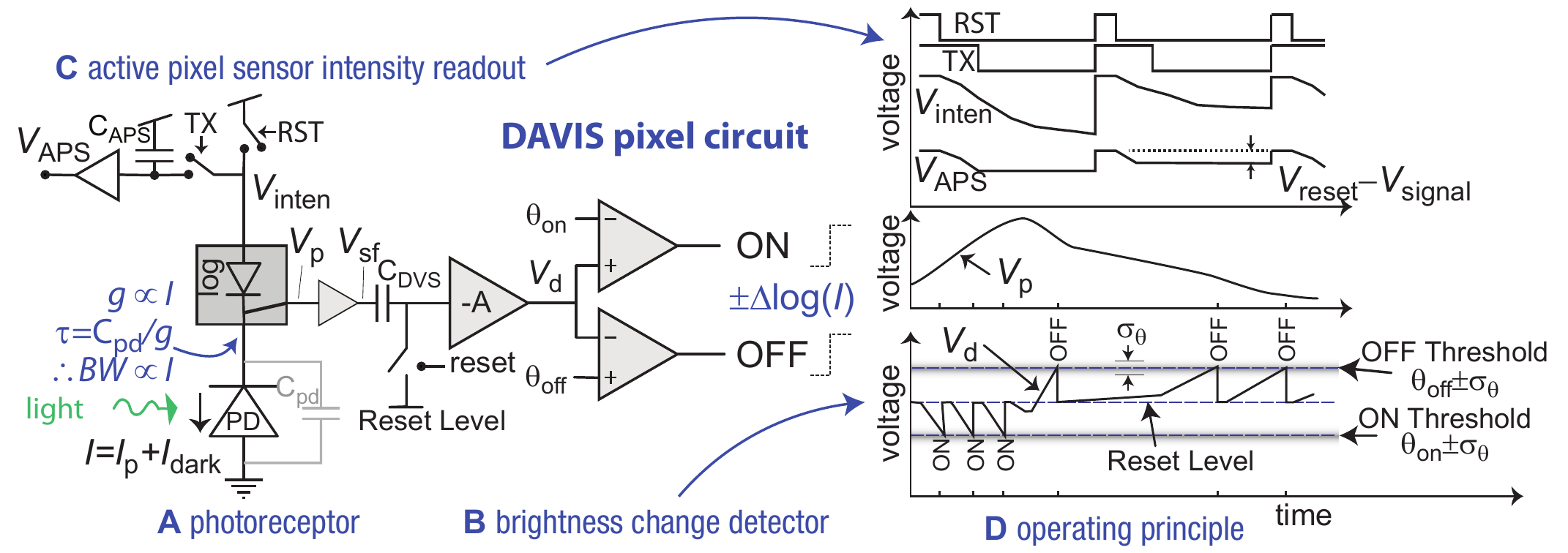}
\caption{\tip{davis} pixel circuit and operating principle. The sensor generates asynchronous brightness change ON and OFF \tip{dvs} events and \tip{aps} intensity samples.}
\label{fig:davis_pixel}
\end{figure*}

Fig.~\ref{fig:davis_pixel} shows the \tip{pdavis} pixel circuit. The design is based on the \tip{dvs}~\autocite{lichtsteiner08} and the \tip{davis}~\autocite{brandli14} with improvements described in~\textcite{taverni18}. 

For the \tip{dvs} brightness change events, the logarithmic photoreceptor (\textbf{A}) drives a change detector (\textbf{B}) that generates the ON and OFF events (\textbf{D}). 
Pixel photoreceptors continuously transduce the photocurrent $I$ produced by the \tip{pd} to a logarithmic voltage $V_\text{p}$, which results in over 120\,dB dynamic range sensitivity. This logarithmic voltage (called \emph{brightness} here) is buffered by a unity-gain source follower to the voltage $V_\text{sf}$, which is stored in a capacitor $C_\text{DVS}$ inside individual pixels, where it is continuously compared to the new input. If the change  $V_\text{d}$ in log intensity exceeds a critical event threshold, an \textsl{ON} or \textsl{OFF} event is generated, representing an increase or decrease of brightness. 
The event thresholds $\theta_\text{on}$ and $\theta_\text{off}$ are nominally identical for the entire array.
The time interval between individual events is inversely proportional to the derivative of the brightness. When an event is generated, the pixel’s location and the sign of the brightness change are immediately transmitted to an arbiter circuit surrounding the pixel array, then off-chip as a pixel address, and a timestamp is assigned to individual events. The arbiter circuit then resets the pixel’s change detector so that a new event can be generated by the pixel. Events can be read out from \tip{pdavis} at up to rates of about 10\,MHz. The quiescent (noise) event rate is a few kHz.

Non-idealities of the \tip{dvs} part of the pixel include 1) finite response time $\tau$ caused by the intensity-dependent RC time constant of the photoreceptor voltage $V_\text{p}$ as indicated in the photoreceptor circuit (\textbf{A}), 2) pixel-to-pixel mismatch $\sigma_{\theta}$ of the brightness change thresholds $\theta_\text{on}$ and $\theta_\text{off}$ (\textbf{D}), and 3) noise in the output~\autocite{Nozaki2017_dvs_temperature_parasitic,Graca2021-iisw-unraveling-dvs-noise}.
These non-idealities lead to background activity~\autocite{Nozaki2017_dvs_temperature_parasitic,Graca2021-iisw-unraveling-dvs-noise} and \tip{fpn}~\autocite{lichtsteiner08} in the \tip{dvs} responses and finite \tip{dvs} motion blur~\autocite{Hu2021-v2e} but in typical operating conditions the temporal jitter of event timing is less than 1\,ms~\autocite{Hu2021-v2e}.

For the frames, the intensity samples are captured by the \tip{aps} readout circuit (\textbf{C}). The absolute intensity is measured by the photocurrent passing non-destructively through the photoreceptor circuit, where it is integrated onto a capacitor $C_\text{APS}$, whose voltage $V_\text{inten}$ is read out via a source follower transistor as $V_\text{APS}$  similar to conventional CMOS image sensors. At the start of each frame, the global signal \textsl{RST} pulls all pixel $V_\text{inten}$ high.  The reset level of each pixel is read out from $V_\text{APS}$ through column-parallel \tips{adc} (not shown). At the end of integration, \textsl{TX} freezes the sampled  $V_\text{inten}$ signal on $C_\text{APS}$ and the signal values are read out. Each final intensity sample is the difference between the reset level and signal level.
The on-chip column-parallel 10-bit \tips{adc} convert the samples of reset and signal and subtraction is computed in software on the host computer. The frame-based output can generate videos with a desired exposure time -- where all pixels have the same integration (or exposure) time -- down to about 10\,us and up to the frame interval. Readout speed limits the maximum frame rate to about 50\,Hz.

Events and frames are transmitted from the \tip{davis} chip to a host computer over \tip{usb} via a programmable logic chip\autocite{Berner2007-100-dollar}. Each frame pixel sample is 10 bits occupying a (non optimized) 2 bytes. Each event is transmitted using a 16-bit microsecond timestamp and from 2 to 4 bytes address depending on data rate (the \tip{davis} uses Boahen's word-serial \tip{aer} interface \autocite{Boahen2004-aer-word-serial-tx-design}); at high data rate, most events use only 2 bytes for their column address, since events from the same row within a short time interval share the same timestamp.

\clearpage
\newpage

\subsection{PDAVIS Characterization Setup}
\label{sup:setup}

%The characterization bench (Fig~\ref{fig:characterization_bench}) is based on the work of \autocite{moeys16} to characterize \tip{dvs} pixels and was extended to vary the \tip{dop} and the \tip{aop}. The light is provided to the integrating sphere from a high-power white LED. The light is then filtered by a rotating polarizer and/or a \tip{qwp} before impinging the surface of the sensor. The rotating polarizer in conjunction with the \tip{qwp}, allows us to vary the \tip{dop} and the \tip{aop} independently from the light intensity. When there is no \tip{qwp} (Fig~\ref{fig:fig2_aop_reconstruction}), a single rotation of the polarizer generates two \tip{aop} cycles. When the \tip{qwp} rotates and the polarizer is fixed, a rotation  generates four \tip{dop} cycles.  The speed of the motor is constant during one measurement, and we assumed that the alignment errors of the setup were negligible compared to the reconstruction errors of the sensors. 

\begin{figure*}[!ht]
\centering
\includegraphics[width=0.8\columnwidth]{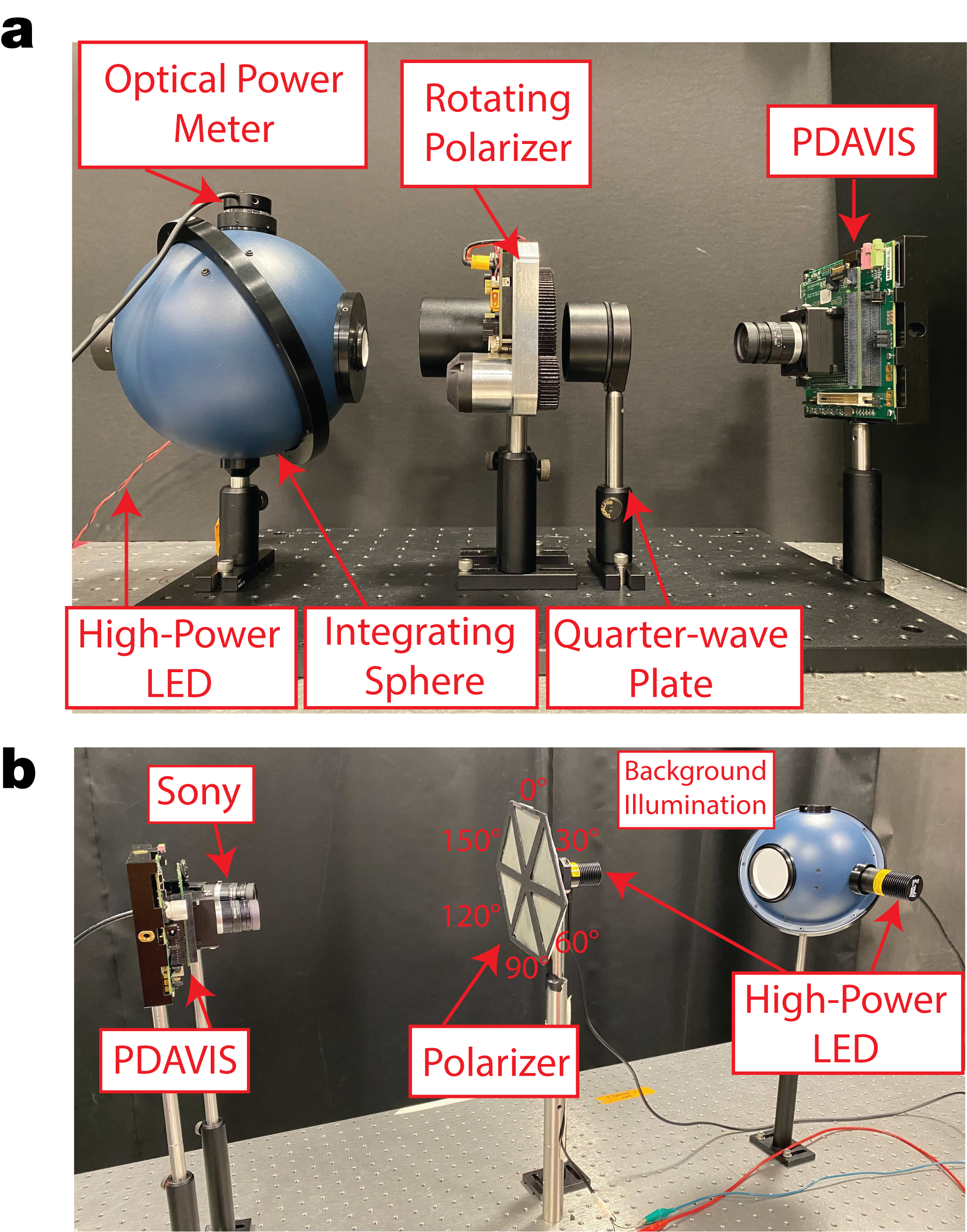}
\caption{\textbf{a} Photo of the characterization bench, built around the integrating sphere to generate an unpolarized light, filtered through either a linear polarizer or a combination of linear polarizer and a \tip{qwp} to generate different \tip{aop} and \tip{dop} profiles while an optical power meter measures the irradiance level. Data is collected with either Sony or \tip{pdavis} sensor. 
\textbf{b} Experimental setup used for collecting data under high dynamic range and rotating ensemble of six linear polarization filters offset by 30 \textdegree.}
\label{fig:characterization_bench}
\end{figure*}

Fig.~\ref{fig:characterization_bench}-a depicts the experimental setup used to evaluate the optoelectronic properties for both PDAVIS and Sony polarization cameras when rotating a linear polarization filter at different speeds. A narrow band LED light source (LZ4-00G108, Osram) centered at 520 nm is coupled to a 6” integrating sphere (819D-SF-5.3, Newport). The light exiting the integrating sphere is uniform and depolarized due to the multiple scattering events inside the integrating sphere. A custom-built rotational stage is placed in front of the output port of the integrating sphere. The rotational stage is controlled via a 10:1 stepper gear and a DC motor with a feedback controller. The rotational speed of the stage is controlled from a computer and can rotate up to 3,000 RPM. Due to the feedback controller, the rotational speed is within 10 RPM of the desired value.

For the first set of experiments, a linear polarization filter (20LP-VIS-B, Newport) is placed in the rotational stage opening. The light emerging from the rotating filter is imaged with either PDAVIS or Sony’s polarization camera. When a linear polarization filter is rotated in front of the camera, the angle of polarization is varied between 0 and 180 degrees. One full rotation of the linear polarization filter will generate two full cycles of the \tip{aop}. The rotational speed of the filter was varied between 60 RPM and 1,000 RPM.

For the second set of experiments, a zero-order \tip{qwp} filter (20RP34-532, Newport) is placed after the rotational stage housing the linear polarization filter and before the imaging sensor. The orientation of the \tip{qwp} is fixed while the linear polarization filter is rotated at constant speed. When the linear polarization filter is rotated in front of the camera, both the angle and degree of linear polarization are varied. One full rotation of the linear polarization filter will generate four full cycles of the \tip{aop} and \tip{dop}.

Fig.~\ref{fig:characterization_bench}-b depicts the experimental setup used to evaluate the high dynamic range and motion blur for both PDAVIS and Sony polarization cameras when rotating an ensemble of six linear polarization filter offset by 30 \textdegree\ at different speeds. A narrow band LED light source (LZ4-00G108, Osram) centered at 520 nm is coupled to a 6” integrating sphere (819D-SF-5.3, Newport). The output light from the integrating sphere is incident on the ensemble of linear polarization filters. A high intensity LED is placed behind the filter wheel to provide additional light intensity. The filter wheel is mounted on a motor such that the filters are rotated at different speeds. Images are collected with both Sony and PDAVIS sensors.

\newpage

\subsection{Fabrication of PDAVIS}
\label{sup:pfa_assembly}

The pixelated \tips{pfa} were fabricated on a quartz substrate by Moxtek Inc. The \tip{pfa} contains four sets of pixels with linear polarization filters offset by 45 degrees. The pixel pitch of the filter array is 18.5 microns and matches the pitch of the pixels in the DAVIS vision sensor. Pixels are isolated by a 2\,um-wide metal shield. 

\begin{figure*}[!h]
\centering
\includegraphics[width=0.95\columnwidth]{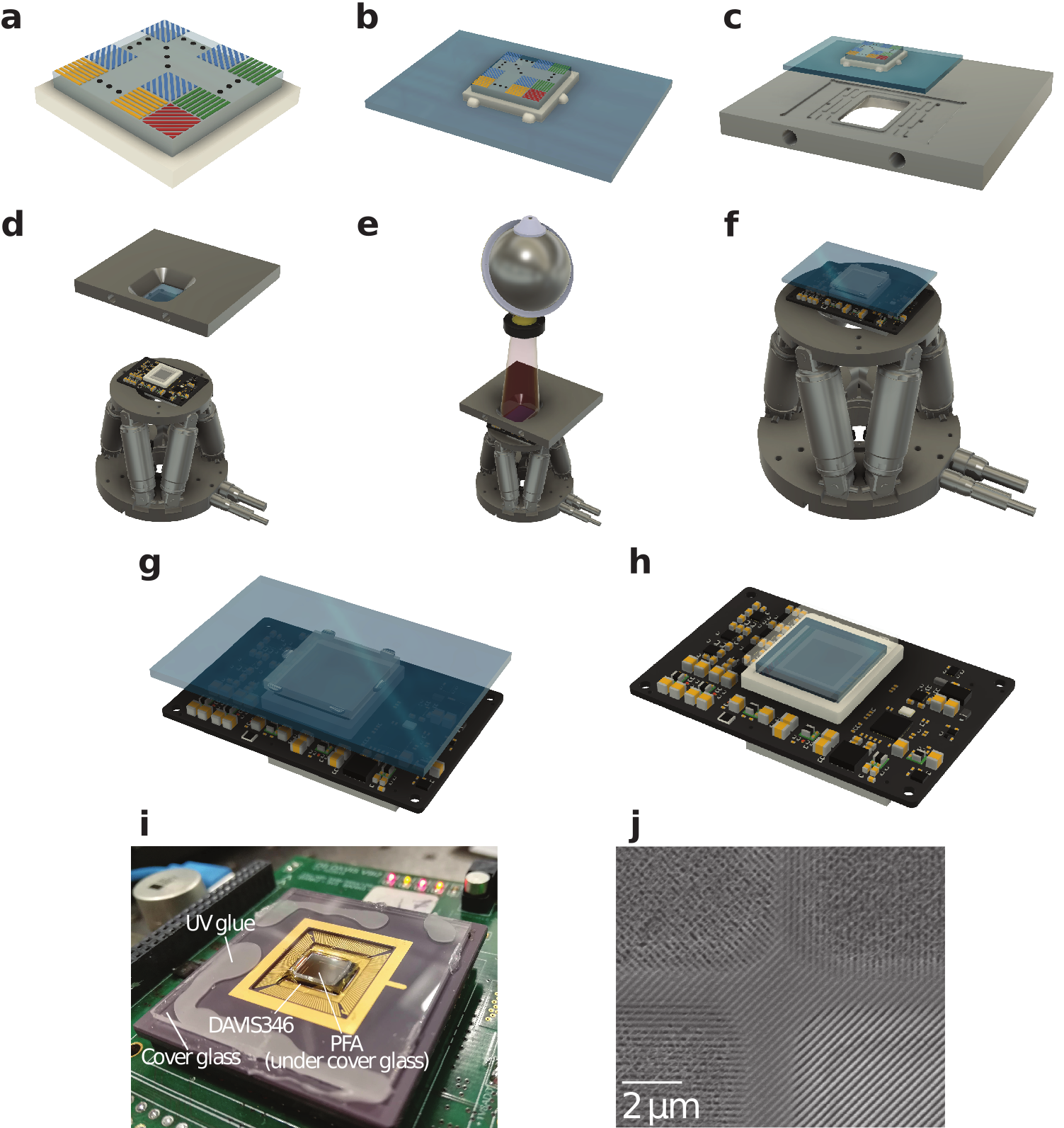}
\caption{%
 \textbf{Fabrication steps of the \tip{pdavis} \tip{pfa}.}
\textbf{a} Quartz glass with \tip{pfa} is glued to a cover glass.
\textbf{b} Cover glass is mounted on a glass plate.
\textbf{c} Glass plate is vacuum-fixed to flat stage,
\textbf{d and e} The polarization filter stack is positioned during live operation of the \tip{pdavis} to the die.
\textbf{f} Polarization filter stack is aligned and attached to the vision sensor.
\textbf{g} Complete image sensor removed from alignment stage.
\textbf{h} Glass plate is removed from the polarization sensor
\textbf{i} Photo of the assembled \tip{pdavis}. 
The \tip{pfa} is attached to the bottom of the coverglass.
\textbf{j} SEM image of \tip{pfa}.}
\label{fig:pfa_assembly}
\end{figure*}

The steps (illustrated in Fig.~\ref{fig:pfa_assembly}) to integrate the \tip{pfa} with the packaged \tip{davis} chip (\ref{sup:davis}) are based on a method initially presented by Blair et al. ~\autocite{Blaireaaw7067} and expanded upon below:

1.	The quartz glass with the \tip{pfa} is glued to a secondary cover glass using a UV activated and optically transparent epoxy (OP-29, Dymax). The cover glass has the same dimensions as the ceramic package of the \tip{davis} (Fig.~\ref{fig:pfa_assembly}a).

2.	The cover glass is mounted on a 3” by 3” glass plate using thermally activated bonding wax (part number) (Fig.~\ref{fig:pfa_assembly}b).

3.	The glass plate is fixated on a custom-built flat stage and held via vacuum on the stage. The entire stage is fixed using 1” posts on a vibration damped optical table (Fig.~\ref{fig:pfa_assembly}c).

4.	The DAVIS vision sensor is mounted on a 6-\tip{dof} alignment stage and placed under the cover glass stage (H-811 Hexapod, PI-USA Instruments). The alignment stage is initialized to the lowest position with about 2 cm space between the filter array and the sensor plane (Fig.~\ref{fig:pfa_assembly}d).

5.	An integrating sphere (819D-SF-4, Newport) coupled with a high power red led (M660D2, Thorlabs) is placed 2 feet away from the DAVIS sensor. An adjustable iris (SM2D25, Thorlabs) followed by a linear polarization filter (WP25L-VIS, Thorlabs) mounted on a computer-controlled rotational stage (HDR50, Thorlabs) is placed at the output of the integrating sphere. The center of the output port of the integrating sphere is aligned with the center of the vision sensor. Due to the large distance between the integrating sphere and the vision sensor and the small aperture of the iris, the incident light is collimated and linearly polarized (Fig.~\ref{fig:pfa_assembly}e).

6.	Live images are streamed from the DAVIS sensor and custom software is used to provide statistics about the vision sensor, such as the mean and standard deviation of individual pixels in the image array as well as extinction ratios between orthogonal pixels.

7.	As the vision sensor is brought closer to the filter array, first the pitch and yaw are adjusted, followed by x- and y- adjustments. For each positional adjustment, the linear polarization filter in front of the vision sensor is rotated, which enables evaluation of the extinction ratios (Fig.~\ref{fig:pfa_assembly}f).

8.	The vision sensor is considered in contact with the pixel array when the extinction ratios remain constant. At this point, the aperture is increased to generate less collimated light. If the extinction ratios remain the same, then the filter is in contact with the sensor.

9.	The DAVIS sensor is then lowered away from the filter array. A UV activated epoxy is added to the ceramic package of the DAVIS sensor and the sensor is brought slowly into contact with the filter array. Because the camera was only translated in the z direction during this step, it remains in good alignment with the filter array. Small positional adjustments can be made if necessary. Since the light source used for imaging does not have any UV component, the epoxy does not cure during the alignment process.

10.	A UV light source is used to activate the epoxy and permanently attach the filters to the ceramic package. The UV light will cure 99\% of the epoxy within the first 15 to 30 seconds. However, 100\% of the epoxy is cured after 24 hours of UV exposure. The filters and vision sensors remain in the alignment stage for 24 hours under UV light to completely cure the epoxy.

11.	Next, the vacuum is turned off so that the glass plate can be removed from the alignment stage without damaging the vision sensor and glued filters. The camera is lowered and removed from the alignment instrument (Fig.~\ref{fig:pfa_assembly}g).

12.	The glass plate is removed by applying heat from a hot plate at 80 C. Any residual bonding wax on top of the vision sensor is cleaned with acetone-soaked cotton swabs (Fig.~\ref{fig:pfa_assembly}h).

The completed \tip{pdavis} is shown in Fig.~\ref{fig:pfa_assembly}i. The pixelated polarization filters are aligned and in contact with the DAVIS pixel array. The filters are attached to a secondary cover glass slide and glued to the ceramic package of the DAVIS chip. 

Fig.~\ref{fig:pfa_assembly}j shows an SEM image of the four different orientations of the nanowires within individual pixelated polarization filters.

\clearpage
\newpage

\subsection{Extinction ratio measurements}
\label{sup:er_measurements}

The following setup is constructed to measure the \tip{er} of PDAVIS as a function of wavelength. We start by using a 400 W halogen light bulb (7787XHP, Philips) housed in a custom box and powered by a high power voltage supply (N5770A, Agilent). The light from the halogen bulb is directed into a monochromator (Acton SP2150, Princeton Instruments), where the desired wavelength is selected using a grating and a computer-controlled slit at the output port. Since the output light beam is typically partially polarized, an integrating sphere (819D-SF-4, Newport) is placed at the output port of the monochromator to depolarize the monochromatic light beam. A pinhole (SM2D25, Thorlabs) followed by an aspherical lens  (ACL5040U, Thorlabs) are used to collimate the light beam emerging from the integrating sphere. Lastly, a linear polarization filter (WP25L-VIS, Thorlabs) mounted on a computer-controlled rotational stage (HDR50, Thorlabs) is used to produce collimated and linearly polarised monochromatic light which is imaged by either the PDAVIS sensor or a spectrometer. The spectrometer is used to measure the exact wavelength of the incident light on the vision sensor. An optical power meter is placed at the same location where the image sensor is located and used to measure the photon flux of the incident light at the desired wavelength.

\begin{figure*}[!h]
\centering
\includegraphics[width=0.8\columnwidth]{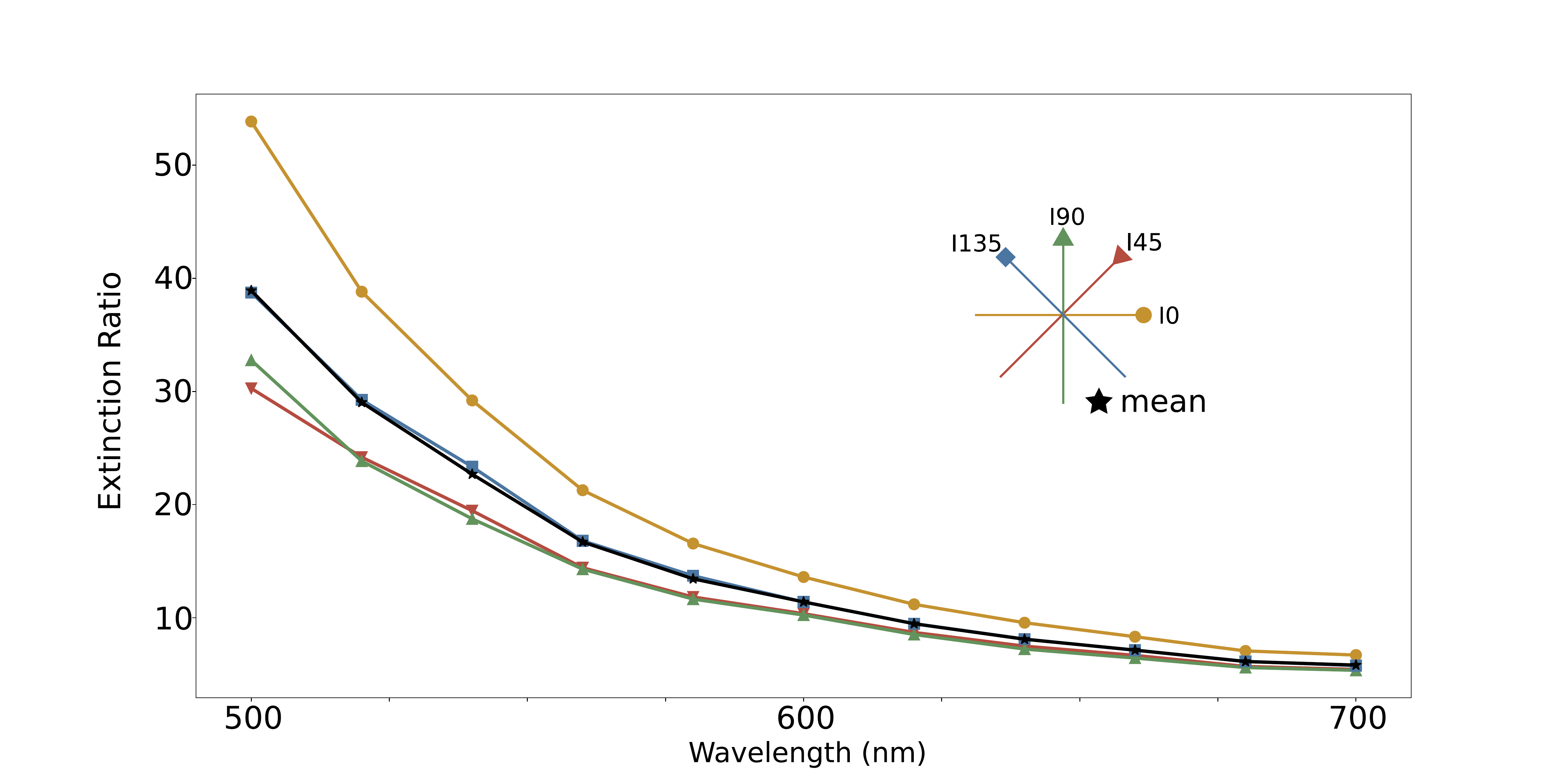}
\caption{%
Measured extinction ratios vs. wavelength for the PDAVIS sensor.
}
\label{fig:er_plot}
\end{figure*}

With the setup described above, we first recorded 100 frames with all light blocked from the sensor. These 100 frames, subsequently called dark frames, are temporally averaged to find the digital number offset for each pixel caused by imperfections of the pixel's circuitry. Next, the monochromator is turned on and set to output monochromatic light at a particular wavelength. The linear polarization filter is rotated 180° in increments of 5°. A total of 100 frames are recorded for each angle and are spatiotemporally averaged over a 42x28 macropixel \tip{roi}. The 180° sweep over a single wavelength gives us one period of Malus’s Law (\ref{sup:conv_pola_imaging}, Eq.~\eqref{eq:fi_vs_dop_and_theta}).
A non-linear regression then fits the cosine squared signal reconstructed from the linear polarizer sweep. After subtracting the dark frame to remove noise offsets, the \tip{er} is computed by taking the ratio of the largest and smallest digital number from the cosine squared function. This process is repeated over the range of wavelengths from 500nm to 700nm in steps of 20nm to give us the extinction ratio for each filter in the macropixel. The results are shown in Figure ~\ref{fig:er_plot}.

\newpage

\subsection{Dynamic range for polarization sensitivity}
\label{sup:dr_measurements}

To measure the \tip{dr} over which the PDAVIS or Sony sensors are sensitive to linearly polarized light, we constructed the following optical setup. A custom-built high power, water-cooled white LED light source (CXB3590, Digikey) is coupled to a 4” integrating sphere (819D-SF-4, Newport). The LED power is adjusted by a computer-controlled power supply (N5770A, Agilent). The light exiting the integrating sphere is uniform and depolarized due to the multiple scattering events inside the integrating sphere. A linear polarization filter (WP25L-VIS, Thorlabs) mounted on a computer-controlled rotational stage (HDR50, Thorlabs) is placed at the output of the integrating sphere. The linear polarization filter is rotated at 60 RPM. Hence, time-varying linearly polarized light at different intensity levels is used to illuminate either the PDAVIS or Sony polarization camera.

The Sony camera exposure time is set to 20\,ms for this set of optical experiments. This mimics a situation where in order for the dark parts of the imaged scene to have a nonzero digital value, the minimum exposure time should be at least 20\,ms. The \tip{pdavis} pixels has a local gain/exposure control by virtue of the logarithmic photoreceptors, and thus there is no need to set any exposure time because the frames are not used for the \tip{dnn} reconstruction method. For each illumination levels, both Sony and \tip{pdavis} capture polarization information from a \tip{roi} of 20x20 macropixel corresponding to the rotating linear polarization filter. For the Sony camera, \tip{aop} is computed from the raw intensity information from the four super pixels and spatially averaged across the \tip{roi}. For the \tip{pdavis}, we first reconstruct the intensity for the four individual channels of polarized pixels using the \tip{dnn} method (which uses only the brightness change events) (see \ref{sup:firenet}), and then computed the spatially averaged \tip{aop} within the \tip{roi}. Since the linear polarization filter was rotating at a steady speed, the \tip{aop} is a repeated sawtooth linear response, sweeping between 0\textdegree\ and 180\textdegree. However, once the illumination level exceeds the \tip{dr} of the camera, the saturated pixels will result in incorrect \tip{aop} computation and thus the range of the reconstructed \tip{aop} is smaller than the expected 180\textdegree.

\begin{figure*}[!h]
\centering
\includegraphics[width=0.7\columnwidth]{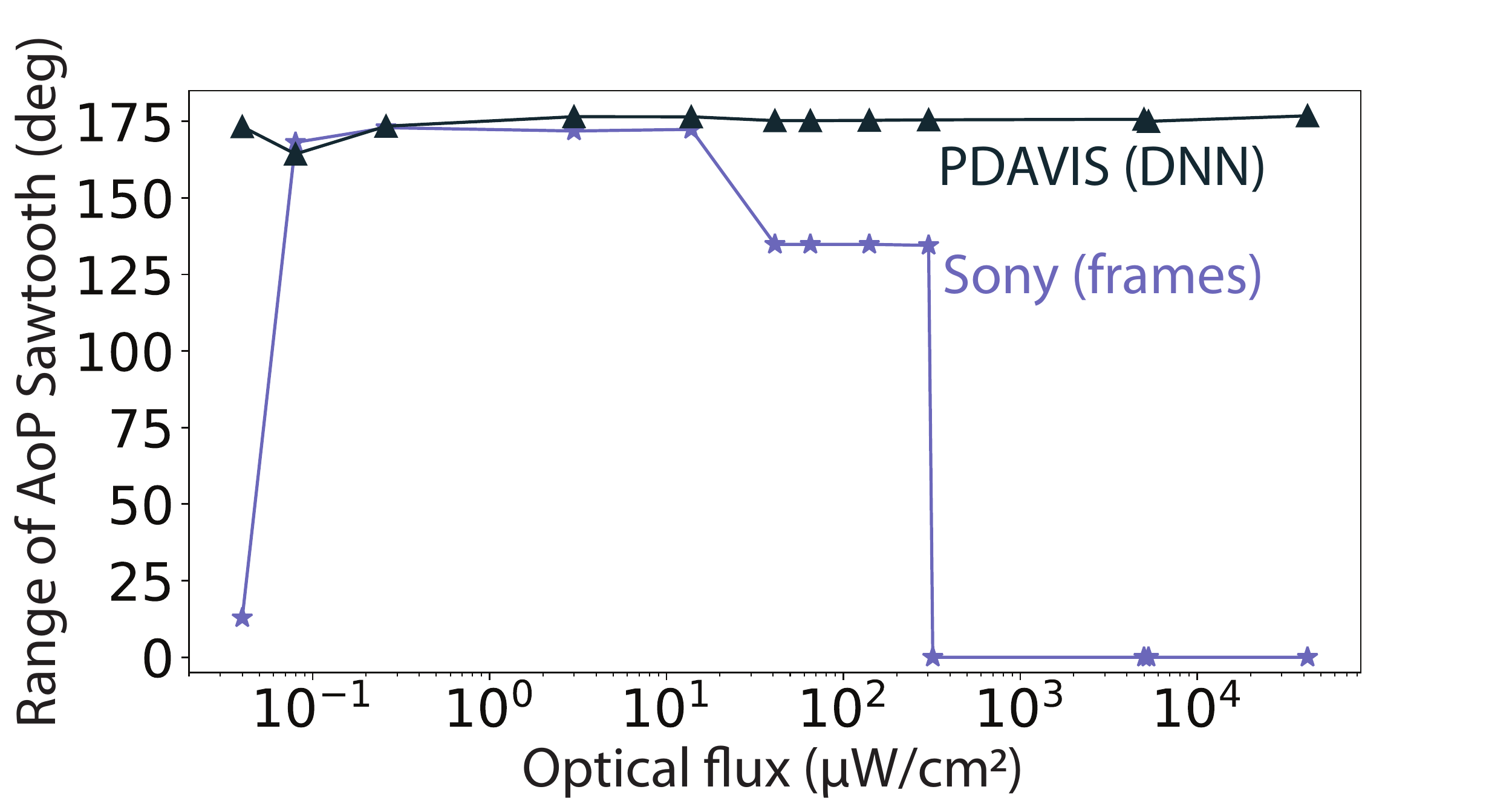}
\caption{%
Measured dynamic range for polarization sensitivity for PDAVIS and Sony's polarization sensors . The \tip{aop} axis is the amplitude of the reconstructed sawtooth as the linear polarization filter is rotated 180 degrees.
}
\label{fig:dr_plot}
\end{figure*}

The Fig.~\ref{fig:dr_plot} results show the range of \tip{aop} reconstruction angles for the sawtooth variation, over illumination level ranging from 40\,nW/cm$^2$ to 42\,mW/cm$^2$. It demonstrates that the Sony camera can reconstruct the \tip{aop} between 40\,nW/cm$^2$ and 300\,$\mu$W/cm$^2$ optical flux, corresponding to a dynamic range of 77.5\,dB (7300X).  Beyond 300\,$\mu$W/cm$^2$ optical flux, all Sony pixels are saturated and the polarization sensitivity vanishes. By contrast, the \tip{pdavis} is able to reconstruct nearly the full 180\textdegree\ sawtooth over 120\,dB illumination levels, from 40\,nW/cm$^2$ to 42\,mW/cm$^2$ optical power.

\newpage

\subsection{PDAVIS and Sony camera specifications}
\label{sup:specifications}

Table~\ref{tab:spec_table} compares the design and measured specifications of the \tip{pdavis} with a state-of-the-art \tip{cots} frame-based polarization camera (FLIR BFS-U3-51S5), which uses the Sony IMX250 camera chip. Details of our measurements of dynamic range and extinction ratio of \tip{pdavis} precede this section.

The Sony polarization sensor has higher resolution, smaller pixel size, and higher extinction ratio. Our bioinspired sensor is fabricated at several different locations: the event-based sensor is fabricated in a 180nm CIS process provided by TowerJazz Semiconductors; pixelated polarization filters are fabricated in Moxtek cleanroom facilities; the filters and image sensor are integrated at University of Illinois. Due to the complex fabrication steps, the image sensor pixel pitch is larger and the extinction ratios are lower than Sony's sensor.
The \tip{pdavis} offers much higher temporal resolution ($\approx$100\,us versus 12\,ms) and its DVS output has superior \tip{dr} compared to Sony polarization camera (120dB vs 72dB).

% \begingroup
% \setlength{\tabcolsep}{4pt} % Default value: 6pt
% \renewcommand{\arraystretch}{1} % Default value: 1

\begin{table}[!h]
\centering
\begin{threeparttable}[!h]
    \caption{Specification and comparison}
    \begin{tabular}{l|c|c}
    & \tip{pdavis} (this work) & \tip{cots} Sony IMX-250\\
    \hline
    \tip{cis} Process & 180nm & 90/40nm stacked \\
    Pixel size & 18.5$\mu$m & 3.45$\mu$m  \\
    %Pixel complexity (FET/Cap) &  47/2 & ??  \\
   Array size  & 346x260 & 2448x2048 \\
   %Photodiode & Buried & HAD??  \\
   %Fill factor & 22\% & ??  \\
    Output & APS+DVS+IMU  & APS \\
    \hline
    %Power consumption (die) &  est 30\,mW & 3W  \\
    Power consumption (camera) &  est. 3\,W & 3\,W  \\
    %Peak QE (sans polarizers) & 24\% & NA \\
    \tipshort{er} at 500nm & 40 & 350 \\
    % \hline
    Max APS frame rate & 53\,Hz\tnote{a} & 75\,Hz  \\
    APS DR & 52dB~\autocite{brandli14} & 72 dB \\
    %APS FPN & 800$\pm$24 DN (3\%) & ?? \\
    %APS read noise & 200e- & NA \\ 
    %APS dark signal & 1200e-/s & NA \\ 
    \hline
    Max DVS event rate & 10 MHz & - \\ 
    DVS DR & 120dB~\autocite{brandli14} & - \\
    DVS Min latency & 3us@1klux~\autocite{brandli14} & - \\
    Min DVS threshold\tnote{b}& $\pm$14\% & - \\ 
    DVS threshold mismatch\tnote{c} & 3.5\%~\autocite{brandli14} & - \\ 
    \end{tabular}
  \begin{tablenotes}
    % \item Best for each metric in \textbf{bold}. 
    \item[a] with exposure 80\,us.
    \item[b] At room temperature, with mean background leak activity rate of 0.7\,Hz with background intensity from APS exposure of 26\,DN/ms.
    \item[c] Pixel to pixel 1-$\sigma$ mismatch of the threshold in temporal contrast.
  \end{tablenotes}
\label{tab:spec_table}
\end{threeparttable}
\end{table}
% \endgroup
\newpage

\section{Reconstructing polarization information from PDAVIS output}
\label{sup:reconstruction_from_pdavis}
\ref{sup:conv_pola_imaging} provides the definition of the Stokes parameters, \tip{aop}, and \tip{dop}.
\ref{sup:event_reconstruction} shows that using only the brightness changes (log intensity changes) signaled by the \tip{dvs} events from \tip{pdavis} allows us to retrieve (at least) the change in \tip{aop}, because it depends on the ratio of differences of polarizer responses. 
However, for the \tip{dop} the absolute intensity does not cancel, and so it cannot be recovered solely from the brightness change events.
The DAVIS \tip{aps} frames, however, provide periodic absolute intensity samples. 
These samples have the limited \tip{dr} and sample rate of frames, but by combining them with the \tip{dvs} events, we can reconstruct the absolute intensity with larger \tip{dr} and higher sample rate, and thus polarization at high effective sample rates. 
We demonstrate two very different approaches for such absolute intensity reconstruction. 
The \tip{cf} (\ref{sup:cf}) is based on a hand-crafted sensory fusion algorithm \tip{cf}, which fuses frames and events. 
The \tip{dnn} method (\ref{sup:firenet}) is based on \tips{dnn} and infers intensity frames from brightness change events, based on its training data.
Table~\ref{tab:comparing_reconstruction} compares the methods.

For Figs.~\ref{fig:fig2_aop_reconstruction}-\ref{fig:fig3_dop_reconstruction}, the plotted \tip{aop} and \tip{dop} values are obtained by averaging over a $12\times 12$ pixel \tip{roi} centered on the rotating polarizer.

\begin{table}[!h]
\caption{Comparison of approaches to reconstruct polarization information from the PDAVIS.}
\label{tab:comparing_reconstruction}
\begin{center}
\begin{tabular}{l|c|c|c|c|p{2cm}|p{2cm}|c}
% \diagbox[width=11em]{Approach}{Properties}
    & \tip{dvs} & \tip{aps} & \tip{aop} & \tip{dop} & \centering Sampling \newline rate [Hz] &  \centering Latency & Op/pixel \\ \hline
Events & \checkmark & -  & \checkmark & -  & \centering 10k   & \centering 1 Event & 12\\
Frames & -  & \checkmark & \checkmark & \checkmark & \centering 25 & \centering 1 Frame & 8\\
\tipshort{dnn} & \checkmark & -  & \checkmark & \checkmark & \centering 10k&  \centering 2k Events & 41k \\
\tipshort{cf} & \checkmark & \checkmark & \checkmark & \checkmark & \centering 10k  & \centering 1 Event or \newline 1 Frame   & 14
\end{tabular}
\end{center}
\end{table}
\newpage

\subsection{Conventional Polarization Imaging}
\label{sup:conv_pola_imaging}

Polarization is commonly described using the Stokes parameters: $(S_0 , S_1 , S_2)$ defined~\autocite{Collett2005-field-guide-to-polarization-2005-spie}:

\begin{equation} \label{eq:s0}
    S_0(t)=I_0(t)+ I_{90}(t)
\end{equation}
\begin{equation} \label{eq:s1}
    S_1(t)=I_0(t)- I_{90}(t)
\end{equation}
\begin{equation} \label{eq:s2}
  S_2(t)=I_{45}(t)- I_{135}(t)    
\end{equation}
\noindent where $I_i$ stands for the light intensity transmitted by the linear polarizer filter with angle $i$. A fourth Stokes parameter ($S3$) describes the circular polarization properties of the light field, which is not detected by the \tip{pdavis} or Sony cameras. 
The \tip{dop} and the \tip{aop} can then be estimated from the Stokes parameters:
\begin{equation}
    \text{DoLP}(t) =  \frac{\sqrt{S_1(t)^2 + S_2(t)^2}}{S_0(t)}  
    \label{eq:dop_stokes}
\end{equation}
\begin{equation}
    \text{AoP}(t)=\frac{1}{2}\arctan\left(\frac{S_2(t)}{S_1(t)} \right)
    \label{eq:aop_stokes}
\end{equation}

Equivalently, the incident light can be separated into unpolarized and linearly polarized beams, whose fluxes are $I_\text{np}$ and $I_\text{p}$, respectively. 
We also let the \tip{aop} be denoted by $\theta(t)$ and the total flux received by the photodiode by $I_\text{t}=I_\text{np}+I_\text{p}$. Thus, the \tip{dop} and the \tip{aop} can be defined by:
\begin{equation}
    \text{DoLP}(t) \equiv \frac{I_\text{p}(t)}{I_\text{t}(t)}
    \label{eq:dop_light}
\end{equation}
\begin{equation}
    \text{AoP}(t)  \equiv \theta(t)
    \label{eq:aop_light}
\end{equation}

Since the intensity is proportional to the square of the electric field strength, the intensity of light passing through each polarizer is given by \eqref{eq:fi_vs_dop_and_theta}~\autocite{Collett2005-field-guide-to-polarization-2005-spie}:
% \footnote{The intensity of a beam of plane-polarized light after passing through a linear polarizer varies as the square of the cosine of the angle through which the polarizer is rotated from the position that gives maximum intensity.} we compute the average intensity passing through each polarizer: 
\begin{equation}
    I_i(t) = I_t(t)   \text{DoLP}(t) \cos[\theta(t) - i]^2 + \frac{I_t(t)}{2}(1-\text{DoLP}(t)).
    \label{eq:fi_vs_dop_and_theta}
\end{equation}
\noindent The second term of \eqref{eq:fi_vs_dop_and_theta} has the factor $\frac{1}{2}$ since half of unpolarized light passes through a filter.
\newpage

% {\color{red}
\subsection{Reconstructing AoP from PDAVIS events only}
\label{sup:event_reconstruction}

We can reconstruct the change in absolute log intensity $dL$ from an arbitrary starting point by simply integrating the events over time, as first studied experimentally by \textcite{Brandli2014-reconstruction}. Pixel nonidealities cause this estimate to drift.
The \textit{events} method\autocite{Scheerlinck2019-complementary-filter} regards the events as providing high frequency information about the log intensity change. Above a corner frequency $\fcorner=2\pi \omega=1/(2\pi \tau)$, the events directly update the filtered log intensity estimate, which decays to zero with time constant $\tau$ between events.
Since the \tip{aop} depends only on ratios of differences of $I_i$ values, the absolute intensity factors out, so we can compute \tip{aop} from the reconstructed $I_i$ values.

For every incoming event, the \textit{events} method asynchronously updates the related  reconstructed log intensity change $dL=d\log(F)$ as the asynchronous first-order \tip{iir} filter \eqref{eq:events_iir}:

\begin{equation}
\label{eq:events_iir}
\begin{split}
    \alpha & \leftarrow e^{-\Delta t/\tau} \\
    dL & \leftarrow\alpha dL+p % \hat{L} & \leftarrow\alpha \hat{L}+p \\ 
    \end{split}
\end{equation}

\noindent where $\Delta t$ is the time elapsed since last event from the subpixel, $\tau$ is the filter time constant, and $p=[+\theta_\text{on},-\theta_\text{off}]$ is the signed event threshold, which we estimate from the known bias currents using the formulas from \textcite{Nozaki2017_dvs_temperature_parasitic} and then fine tune to match the low frequency frame-based data.  

From the $dL$ values, we can compute \tip{aop} by exponentiation of $dL$ to obtain the subpixel $I_i$ value, and use the resulting $I_i$ values in \eqref{eq:aop_stokes}. 
In practice, we use the $dL$ values directly, since generally $|dL|<1$ and thus $\exp(dL)\approx 1+dL$. The 1 would be the same for all terms in \eqref{eq:aop_stokes} and would thus cancel, leaving the $dL$ value.

The $dL$ in \eqref{eq:events_iir} is the highpass-filtered log intensity, corresponding to the Laplace domain transfer function \eqref{eq:events_xfer_function}:
\begin{equation}
    \label{eq:events_xfer_function}
    H_\text{dL}(s)\equiv \frac{dL(s)}{\sum{p}(s)}=\frac{\tau s}{1+\tau s}
    % \hat{L}(s) = \frac{\tau}{a + \tau s} dL(s) = \frac{\tau s}{1 + \tau s}L(s)
\end{equation}
\noindent where $s$ is the complex frequency, and $\sum{p}$ is the staircase sum of Dirac delta brightness changes since filter startup.
% Equivalently, we can observe that 
% \begin{equation}
%     \label{eq:events_deriv_xfer_function}
%     \frac{H_\text{dL}(s)}{\tau s}=\frac{1}{1+\tau s}.
% \end{equation}

There are two exactly equivalent descriptions of this filter: $dL$ is a highpass-filtered log intensity, and it is also a lowpass-filtered derivative of log intensity. Thus, for frequencies well below the $\fcorner$ corner frequency  $dL\approx \tau dL/dt$, \ie $dL$ can be considered as a lowpass filtered derivative of $L$, which filters out derivatives above $\fcorner$. For frequencies well above $\fcorner$, $dL$ is equal to $L$ minus its DC value averaged over the exponential time window $\tau$. If we can assume that this offset is equivalent for each subpixel, then it cancels out in $S_1$ and $S_2$, which are used to compute the \tip{aop}. 

For the \textit{events} results in Figs.~\ref{fig:fig1_header} and \ref{fig:fig2_aop_reconstruction} we used $\fcorner=0.5\text{ Hz}$.

% Since the events represent a brightness change, we can not  directly apply \eqref{eq:aop_stokes}. Ignoring the imperfections of the \tip{dvs} circuitry, we can nevertheless demonstrate that inserting a temporal derivative of the $I_i$ signals (as described in \eqref{eq:fi_vs_dop_and_theta}) in \eqref{eq:aop_stokes} allows us to retrieve the \tip{aop} information.

\textbf{Effect of high pass filter on \tip{aop}:}
For input frequencies well above $\fcorner$, using the $dL$ values in \eqref{eq:aop_stokes} results in the \tip{aop} if we make the reasonable assumption that all $I_x$ have the same mean value. For frequencies below $\fcorner$, where $dL\approx \tau dL/dt$, the following computation shows that using $dL$ in the \tip{aop} equation \eqref{eq:aop_stokes} results in the \tip{aop}, but with a phase shift of $\pi/4$. First, we use \eqref{eq:fi_vs_dop_and_theta} to compute the derivative of $I_i$, where $i$ is one of the polarizer angles:

\begin{equation}
    \frac{\partial I_i(t)}{\partial t}  =-2 I_t \text{DoLP}(t) \frac{\partial \theta(t)}{\partial t} \sin \left[\theta(t) - i\right] \cos\left[\theta(t) - i\right]
    \label{eq:dF_lin}
\end{equation}

Since we only care about measuring a varying \tip{aop}, we have assumed that \tip{dop} and $I_t$ are constant. Now we can plug \eqref{eq:dF_lin} into \eqref{eq:aop_stokes}:

\begin{equation}
\begin{split}
\frac{1}{2}\arctan\left(\frac{\partial S_2/\partial t}{\partial S_1/\partial t} \right) & =  \\
= & \frac{1}{2}\arctan\left(\frac{\partial I_{45}/\partial t - \partial I_{135}/\partial t}{\partial I_{0}/\partial t-\partial I_{90}/\partial t} \right) \\
= & \frac{1}{2} \arctan\left( \frac{2\sin(\theta - \frac{\pi}{4})\cos(\theta - \frac{\pi}{4}) }{2\sin(\theta)\cos(\theta)  }  \right) \\
= & \frac{1}{2} \arctan\left( \frac{1}{-\tan(2\theta)}  \right) \\
= & \theta(t) \bmod \pi + \frac{\pi}{4} \\
= & \text{AoP}(t) \bmod \pi + \frac{\pi}{4} 
\label{eq:aop_event}
\end{split}
\end{equation}

According to \eqref{eq:aop_event}, using the temporal derivatives of intensities in \eqref{eq:aop_stokes} results in the \tip{aop} with a (constant $\pi / 4$) offset.

In practice, we used signal periodicity to estimate the \tip{aop} phase in Fig.~\ref{fig:fig2_aop_reconstruction}.
Most of our experiments used a stimulus frequency above $\fcorner$, so the output of the \tip{aop} from the events method corresponds to the actual \tip{aop} without this offset.
For example, Fig.~\ref{fig:fig2_aop_reconstruction}c shows the reconstructed \tip{aop} sawtooth at 30\,RPM, corresponding to an \tip{aop} frequency of 1Hz, which is double the $\fcorner=0.5\text{ Hz}$ corner frequency.

To lower the effect of mismatch and temporal noise, an event generated by the macropixel $(x, y)$ contributes to update the macropixels within a one pixel radius. 

The \tip{aop} values are updated as soon as each event is received, creating polarization events as illustrated in Fig.~\ref{fig:fig1_header}c. These asynchronous updates could drive a quick event-driven processing pipeline that exploits the precise timing of events.

Source code for this algorithm is available \footnote{\url{https://github.com/joubertdamien/poladvs}}.

\newpage
\subsection{Complementary Filter: Reconstructing AoP and DoLP by fusing frames and events}
\label{sup:cf}

The \tip{cf} of Scheerlinck\autocite{Scheerlinck2019-complementary-filter} is \textit{complementary} because it considers \tip{aps} frames as providing reliable low frequency intensity (albeit with limited \tip{dr}), while the \tip{dvs} events provide reliable high frequency information about brightness (changes).
The \tip{cf} method  fuses the high pass filtered log intensity of the events method with low pass filtered frames.
At the \tip{cf} crossover frequency $\omega=1/\tau=2\pi \fcorner$, the frame and event estimates of log intensity are weighted equally. For lower frequencies, the frame intensities are weighted more, and for higher frequencies, the event-based estimations are weighted more.

The \tip{cf} also has a computational cost of about 10 operations per \tip{dvs} event or \tip{aps} sample, making it attractive for real-time applications.

At each subpixel, the \tip{cf} updates its log intensity reconstruction $L$ each time the pixel measures either intensity or generates a \tip{dvs} event. The \tip{cf} outputs the log intensity $L$ from the most recent log intensity sample or \tip{dvs} event.
For each pixel's \tip{aps} intensity sample or \tip{dvs} event, the asynchronous first-order \tip{iir} filter \tip{cf} update is
\begin{equation}
\begin{split}
\label{eq:cf}
    \alpha & \leftarrow e^{-\Delta t/\tau} \\ 
    L & \leftarrow \underbrace{\alpha L+p}_\text{DVS}+\underbrace{(1-\alpha) L_\text{aps}}_\text{APS}
\end{split}
\end{equation}

\noindent where $\Delta t$ is the time since last update, $\tau=1/(2\pi \fcorner)$ is the filter time constant, $p=[+\theta_\text{on},-\theta_\text{off}]$ is the event's log intensity change, and $L_\text{aps}$ is the log intensity sample. (If the update is for an event, $L_\text{aps}=0$, or if the update is for a frame, $p=0$.)  
Since $\Delta t\ll\tau$ (\ie the update rate is much higher than the time constant), $\alpha\approx 1-\Delta t/\tau\sim 1$.
% We can observe that if $\Delta t$ is large, then   $L$ approaches the latest $L_\text{aps}$ value, plus the latest event.
Removing the \tip{aps} input from Eq.~\ref{eq:cf} gives the \textit{events} method presented in the previous section (Eq.~\ref{eq:events_iir}).

In the Laplace domain, the \tip{cf} filter has form \eqref{eq:cf_xfer_function}:
\begin{equation}
    \label{eq:cf_xfer_function}
   L(s)=\frac{\tau s}{1+\tau s} \sum{p(s)} + \frac{1}{1+\tau s}L_\text{APS}(s).
\end{equation}

\begin{figure*}[!h]
\centering
\includegraphics[width=0.5\columnwidth]{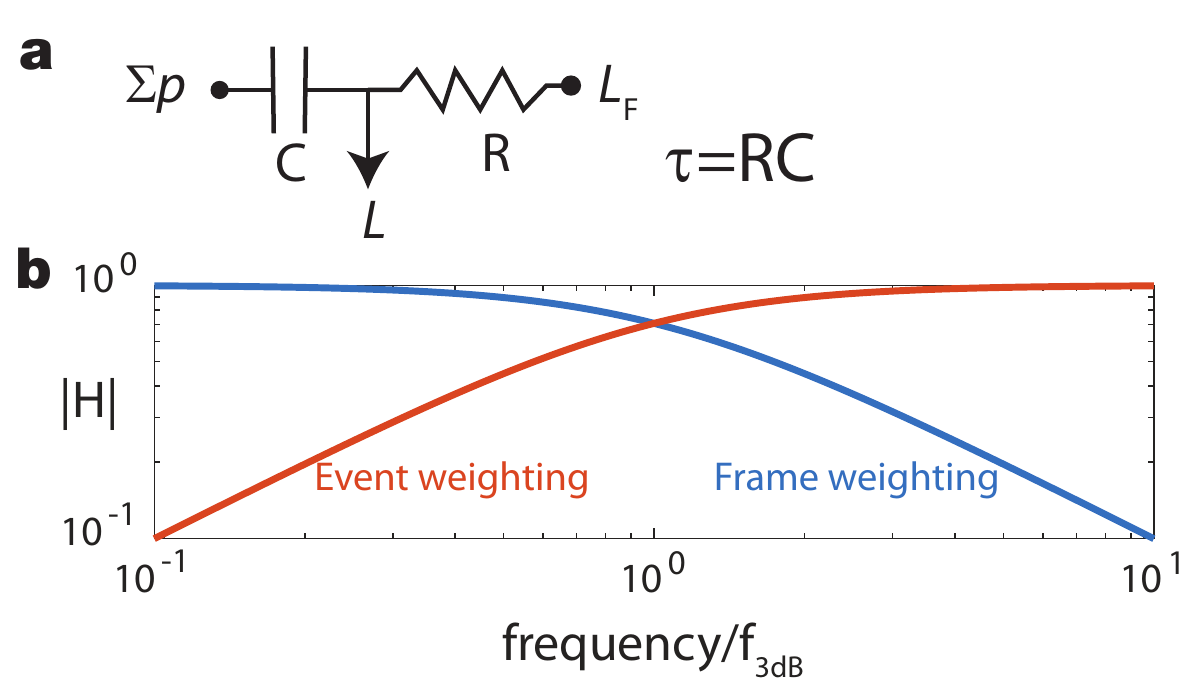}
\caption{%
\textbf{a} Equivalent continuous-time circuit of \tip{cf}. \textbf{b} Magnitude transfer function of the \tip{cf} method.
}
\label{fig:cf_transfer_function}
\end{figure*}

Fig.~\ref{fig:cf_transfer_function} illustrates the equivalent continuous time \textit{CR} highpass plus \textit{RC} lowpass circuit for the \tip{cf}, along with the weighting of frames and events in the resulting \tip{cf} transfer function.

For our experiments, we used \tip{cf} $\fcorner=1.6\text{ Hz}$.  The \tip{aop} and \tip{dop} are periodically computed using the subpixel $L$ values.
The user can decide this update rate; by default, the computation occurs at the end of each packet of \tip{pdavis} data.

\textbf{Adaptive gain tuning}: The \tip{cf} method includes a downweighting of the \tip{aps} samples when $L_\text{aps}$ approach their limits, \ie are under or overexposed~\cite[Sec. 4.1]{Scheerlinck2019-complementary-filter}. We used this feature to improve the \tip{dr} of the reconstruction. We set adaptive gain tuning $\lambda=0.1$ and used the limits $L_1,L_2=\log(10,200)$.

\textbf{Filter startup:} To avoid the \tip{cf} filter startup transient, we initialize the filter output state to the first $L_\text{APS}$ frame as soon as it is available.

% \textbf{Enhancements:} For high speed polarization video, we improved the original \tip{cf} method by using the actual \tip{aps} exposure time (the midpoint of start and end of exposure), rather than the end of readout time (as implemented in code of \autocite{Scheerlinck2019-complementary-filter}), when the \tip{aps} samples are actually available from the camera. To implement this improvement, we added a buffer to hold the \tip{dvs} events staring at the start of exposure until the end of the frame readout. This improved reconstruction comes with the cost of additional latency, since we must wait for the frame sample readout before updating the \tip{cf}. Thus, there is a pause after each start of exposure until the end of the frame readout.

 Source code for the original \tip{cf} implementation, our implementation of \tip{cf}, and for computing \tip{pdavis} polarization information are  open-source\footnote{\href{https://github.com/cedric-scheerlinck/dvs_image_reconstruction}{Original CF implementation of \textcite{Scheerlinck2019-complementary-filter}},  \href{https://github.com/SensorsINI/jaer/blob/master/src/ch/unizh/ini/jaer/projects/davis/frames/DavisComplementaryFilter.java}{\textsf{DavisComplementaryFilter}},  \href{https://github.com/SensorsINI/jaer/blob/master/src/au/edu/wsu/PolarizationComplementaryFilter.java}{\textsf{PolarizationComplmentaryFilter}}, \href{https://github.com/joubertdamien/pyComplementaryFilter}{Fast C++ implementation}}.
\newpage

\subsection{Polarization FireNet: Reconstructing AoP and DoLP from events}
\label{sup:firenet}

The \tip{dnn} method applied to the PDAVIS data is based on deep learning and infers the intensity sensed by each subpixel using only the brightness change events.
It is based on the work of \cite{scheerlinck2020fast,rebecq2019high}, which showed that it is possible to train a deep recurrent neural network to reconstruct video purely from \tip{dvs} brightness change events, as long as there is motion in the scene. 
The reconstructed offset level is chosen by the \tip{dnn} based on the statistics of its training data samples since the \tip{dvs} output transmits no offset information, but the reconstruction is locally more accurate in comparison to the \tip{cf} method. 

We started with a pretrained \textsl{FireNet} \autocite{scheerlinck2020fast} neural network. 
For the polarization reconstruction, the events are first separated into 4 channels, each corresponding to one pixel of 2-by-2 macropixels. Each channel thus represents one out of four different polarization angles (see Sec.~\ref{sup:pdavis_camera} and Fig.~\ref{fig:fig1_header}). 
The events are then accumulated into 3D tensors with the same predetermined exposure time window for each channel, which is
different from the original \textsl{FireNet} which used constant event-count exposures.
This binning requires that the necessary sample rate must be known apriori to obtain a precise reconstruction of the polarization information. 
To synchronize the four channels, we used a fixed time window in opposition to a fixed event count, because each channel codes for different and sometimes even orthogonal angles of polarization, hence emitting a different number of events. For example, in the \tip{dr} measurement, the time window is set to 10\,ms.

Once we receive the stack of frames from the \textsl{FireNet}, calibration is applied. For the data collected using only a linear polarizer (Fig.~\ref{fig:fig2_aop_reconstruction}), we subtract an offset calculated by the minimum value of each of the 4 channels before calculating \tip{aop}. For the \tip{dop} calculation, a gain table of the digital numbers paired with its respective multipliers is made for each RPM from one \tip{aop} cycle. This table gives us the non linear mapping from logarithmic to linear response for each of the four channels that is used on a second data set to calculate the corrected \tip{dop}. As for the \tip{dop} calculation of the data collected from the linear polarizer and quarter wave plate (Fig.~\ref{fig:fig3_dop_reconstruction}), the 4 channels only have an offset and normalization of the max data point applied before \tip{dop} calculation.
Then, the \textsl{FireNet} outputs intensity frames from the event tensors, which we use to compute the angle and degree of polarization using Eqs.~\eqref{eq:aop_stokes} and \eqref{eq:dop_stokes}.

Our source code for  \textsl{FireNet} reconstruction is available on GitHub\footnote{\url{https://github.com/tylerchen007/firenet-pdavis}}.
\newpage

\section{Bovine tendon preparation and experimental setup}
\label{sup:bovine}
Bovine flexor tendon was sliced using a vibratome to produce 300-micron thick slices. A single sliced tendon is mounted on a 6-\tip{dof} computer-controlled actuator and sensor stage. The end pieces of the tendon are clamed via sandpaper to the sensor stage. An LED light source combined with a linear polarization filter (Gray Polarizing Film 38-491, Edmund Optics) and an achromatic \tip{qwp} (AWQP3, Bolder Vision Optik, Boulder, Colorado) were placed under the bovine flexor tendon. This optical setup generates circularly polarized light which is used to illuminate the tendon. The light that is transmitted through the tissue is imaged with either the \tip{pdavis} or Sony polarization sensor. Both sensors were equipped with x10 optical lens with a numerical aperture of 0.25 (f/2) and placed directly above the tissue.

The tendon is cyclically loaded between 2\% and 3\% strain at rates of 1, 5, and 10 Hz for 30 seconds. During the cyclical loading of the tissue, the birefringent properties of the individual collagen fibers are modulated as a function of the applied strain. As circularly polarized light is transmitted through the tissue under cyclical load, the light passing through the collagen fibers will both scatter (\ie depolarized light) and become elliptically polarized. The ellipticity of the polarized light is directly proportional to the strain applied to the collagen fibers. Hence, the degree of linear polarization provides a measurement of the ellipticity of the circularly polarized light and an indirect measure of the applied strain on the tendon.
\newpage

\printbibliography

@article{brandli14,
  title={A 240$\times$180 130 {dB} 3 $\mu$s latency global shutter spatiotemporal vision sensor},
  author={Brandli, Christian and Berner, Raphael and Yang, Minhao and Liu, Shih-Chii and Delbruck, Tobi},
  journal={IEEE Journal of Solid-State Circuits},
  volume={49},
  number={10},
  pages={2333--2341},
  year={2014},
  publisher={IEEE}
}

@article{taverni18,
  title={Front and back illuminated dynamic and active pixel vision sensors comparison},
  author={Taverni, Gemma and Moeys, Diederik Paul and Li, Chenghan and Cavaco, Celso and Motsnyi, Vasyl and Bello, David San Segundo and Delbruck, Tobi},
  journal={IEEE Transactions on Circuits and Systems II: Express Briefs},
  volume={65},
  number={5},
  pages={677--681},
  year={2018},
  publisher={IEEE}
}

@article{gruev10,
  title={{CCD} polarization imaging sensor with aluminum nanowire optical filters},
  author={Gruev, Viktor and Perkins, Rob and York, Timothy},
  journal={Optics express},
  volume={18},
  number={18},
  pages={19087--19094},
  year={2010},
  publisher={Optical Society of America}
}

@article{lichtsteiner08,
  title={A 128x128 120 dB 15us Latency Asynchronous Temporal Contrast Vision Sensor},
  author={Lichtsteiner, Patrick and Posch, Christoph and Delbruck, Tobi},
  journal={IEEE journal of solid-state circuits},
  volume={43},
  number={2},
  pages={566--576},
  year={2008},
  publisher={IEEE}
}

@inproceedings{scheerlinck2020fast,
  title={Fast image reconstruction with an event camera},
  author={Scheerlinck, Cedric and Rebecq, Henri and Gehrig, Daniel and Barnes, Nick and Mahony, Robert and Scaramuzza, Davide},
  booktitle={Proceedings of the IEEE/CVF Winter Conference on Applications of Computer Vision},
  pages={156--163},
  year={2020}
}

@INPROCEEDINGS{Scheerlinck2019-complementary-filter,
  title     = "{Continuous-Time} Intensity Estimation Using Event Cameras",
  booktitle = "Computer Vision -- {ACCV} 2018",
  author    = "Scheerlinck, Cedric and Barnes, Nick and Mahony, Robert",
  publisher = "Springer International Publishing",
  pages     = "308--324",
  year      =  2019,
  url       = "http://dx.doi.org/10.1007/978-3-030-20873-8_20",
  doi       = "10.1007/978-3-030-20873-8\_20"
}

@article{rebecq2019high,
%   title={High speed and high dynamic range video with an event camera},
%   author={Rebecq, Henri and Ranftl, Ren{\'e} and Koltun, Vladlen and Scaramuzza, Davide},
%   journal={IEEE transactions on pattern analysis and machine intelligence},
%   year={2019},
%   publisher={IEEE}
% }

@article{marshall2019polarisation,
  title={Polarisation signals: a new currency for communication},
  author={Marshall, N Justin and Powell, Samuel B and Cronin, Thomas W and Caldwell, Roy L and Johnsen, Sonke and Gruev, Viktor and Chiou, T-H Short and Roberts, Nicholas W and How, Martin J},
  journal={Journal of Experimental Biology},
  volume={222},
  number={3},
  year={2019},
  publisher={The Company of Biologists Ltd}
}

@ARTICLE{Gallego2020-survey-paper,
  title    = "Event-based Vision: A Survey",
  author   = "Gallego, Guillermo and Delbruck, Tobi and Orchard, Garrick
              Michael and Bartolozzi, Chiara and Taba, Brian and Censi, Andrea
              and Leutenegger, Stefan and Davison, Andrew and Conradt, Jorg and
              Daniilidis, Kostas and Scaramuzza, Davide",
  journal  = "IEEE Trans. Pattern Anal. Mach. Intell.",
  volume   = "PP",
  pages    = "1--1",
  month    =  jul,
  year     =  2020,
  url      = "http://dx.doi.org/10.1109/TPAMI.2020.3008413",
  language = "en",
  issn     = "0162-8828",
  pmid     = "32750812",
  doi      = "10.1109/TPAMI.2020.3008413"
}

@BOOK{Collett2005-field-guide-to-polarization-2005-spie,
  title     = "Field guide to polarization",
  author    = "Collett, Edward",
  publisher = "spie.org",
  year      =  2005,
  url       = "https://spie.org/Publications/Book/626141",
  isbn      = "9780819458681"
}

@article{Marshall88,
  title={A unique colour and polarization vision system in mantis shrimps},
  author={Marshall, N Justin},
  journal={Nature},
  volume={333},
  number={6173},
  pages={557--560},
  year={1988},
  publisher={Nature Publishing Group}
}

@article {Altaquieabe3196,
	author = {Altaqui, Ali and Sen, Pratik and Schrickx, Harry and Rech, Jeromy and Lee, Jin-Woo and Escuti, Michael and You, Wei and Kim, Bumjoon J. and Kolbas, Robert and O{\textquoteright}Connor, Brendan T. and Kudenov, Michael},
	title = {Mantis shrimp{\textendash}inspired organic photodetector for simultaneous hyperspectral and polarimetric imaging},
	volume = {7},
	number = {10},
	elocation-id = {eabe3196},
	year = {2021},
	doi = {10.1126/sciadv.abe3196},
	publisher = {American Association for the Advancement of Science},
	URL = {https://advances.sciencemag.org/content/7/10/eabe3196},
	eprint = {https://advances.sciencemag.org/content/7/10/eabe3196.full.pdf},
	journal = {Science Advances}
}

@article{GarciaPolarization,
author = {Missael Garcia and Christopher Edmiston and Radoslav Marinov and Alexander Vail and Viktor Gruev},
journal = {Optica},
number = {10},
pages = {1263--1271},
publisher = {OSA},
title = {Bio-inspired color-polarization imager for real-time in situ imaging},
volume = {4},
month = {10},
year = {2017},
url = {http://www.osapublishing.org/optica/abstract.cfm?URI=optica-4-10-1263},
doi = {10.1364/OPTICA.4.001263},
}

@article{LiuEndoscope,
author = {Chenyang Liu and Chengyong Shi and Taisheng Wang and Hongxin Zhang and Lei Jing and Xiya Jin and Jia Xu and Hongying Wang},
journal = {Opt. Express},
number = {1},
pages = {145--157},
publisher = {OSA},
title = {Bio-inspired multimodal 3D endoscope for image-guided and robotic surgery},
volume = {29},
month = {1},
year = {2021},
url = {http://www.opticsexpress.org/abstract.cfm?URI=oe-29-1-145},
doi = {10.1364/OE.410424},
}

@article{BioinspiredSenorsReview2021,
author = {Kim, Min Sung and Kim, Min Seok and Lee, Gil Ju and Sunwoo, Sung-Hyuk and Chang, Sehui and Song, Young Min and Kim, Dae-Hyeong},
title = {Bio-Inspired Artificial Vision and Neuromorphic Image Processing Devices},
journal = {Advanced Materials Technologies},
volume = {n/a},
number = {n/a},
pages = {2100144},
doi = {https://doi.org/10.1002/admt.202100144},
url = {https://onlinelibrary.wiley.com/doi/abs/10.1002/admt.202100144},
eprint = {https://onlinelibrary.wiley.com/doi/pdf/10.1002/admt.202100144}
}

@article{jen2011biologically,
  title={Biologically inspired achromatic waveplates for visible light},
  author={Jen, Yi-Jun and Lakhtakia, Akhlesh and Yu, Ching-Wei and Lin, Chia-Feng and Lin, Meng-Jie and Wang, Shih-Hao and Lai, Jyun-Rong},
  journal={Nature communications},
  volume={2},
  number={1},
  pages={1--5},
  year={2011},
  publisher={Nature Publishing Group}
}

@article {Blaireaaw7067,
	author = {Blair, Steven and Garcia, Missael and Davis, Tyler and Zhu, Zhongmin and Liang, Zuodong and Konopka, Christian and Kauffman, Kevin and Colanceski, Risto and Ferati, Imran and Kondov, Borislav and Stojanoski, Sinisa and Todorovska, Magdalena Bogdanovska and Dimitrovska, Natasha Toleska and Jakupi, Nexhat and Miladinova, Daniela and Petrusevska, Gordana and Kondov, Goran and Dobrucki, Wawrzyniec Lawrence and Nie, Shuming and Gruev, Viktor},
	title = {Hexachromatic bioinspired camera for image-guided cancer surgery},
	volume = {13},
	number = {592},
	elocation-id = {eaaw7067},
	year = {2021},
	doi = {10.1126/scitranslmed.aaw7067},
	publisher = {American Association for the Advancement of Science},
	issn = {1946-6234},
	URL = {https://stm.sciencemag.org/content/13/592/eaaw7067},
	eprint = {https://stm.sciencemag.org/content/13/592/eaaw7067.full.pdf},
	journal = {Science Translational Medicine}
}

@ARTICLE{Powelleaao6841,
    author  =   "Powell, Samuel B. and Garnett, Roman and Marshall, Justin and Rizk, Charbel and Gruev, Viktor",
	title = "Bioinspired polarization vision enables underwater geolocalization",
	volume = "4",
	number = "4",
	elocation-id = "eaao6841",
	year = "2018",
	doi = "10.1126/sciadv.aao6841",
	publisher = "American Association for the Advancement of Science",
	URL = "https://advances.sciencemag.org/content/4/4/eaao6841",
	eprint = "https://advances.sciencemag.org/content/4/4/eaao6841.full.pdf",
	journal = "Science Advances"
}

@book{cronin2014visual,
  title={Visual ecology},
  author={Cronin, Thomas W and Johnsen, S{\"o}nke and Marshall, N Justin and Warrant, Eric J},
  year={2014},
  publisher={Princeton University Press}
}

@MISC{Sony2021-triton-imx250,
  title        = "Triton {5.0MP} Polarization Camera, Sony's {IMX250MZR} /
                  {MYR} {CMOS}",
  url          = "https://thinklucid.com/product/triton-5-mp-polarization-camera/",
  howpublished = "\url{https://thinklucid.com/product/triton-5-mp-polarization-camera/}",
  note         = "Accessed: 2021-8-13"
}

@INPROCEEDINGS{Hu2021-v2e,
  title     = "V2e: From video frames to realistic {DVS} events",
  booktitle = "Proceedings of the {IEEE/CVF} Conference on Computer Vision and
               Pattern Recognition",
  author    = "Hu, Yuhuang and Liu, Shih-Chii and Delbruck, Tobi",
  pages     = "1312--1321",
  year      =  2021,
  url       = "https://openaccess.thecvf.com/content/CVPR2021W/EventVision/html/Hu_v2e_From_Video_Frames_to_Realistic_DVS_Events_CVPRW_2021_paper.html"
}

@article{PATEL20201981,
title = {Mantis Shrimp Navigate Home Using Celestial and Idiothetic Path Integration},
journal = {Current Biology},
volume = {30},
number = {11},
pages = {1981-1987.e3},
year = {2020},
issn = {0960-9822},
doi = {https://doi.org/10.1016/j.cub.2020.03.023},
url = {https://www.sciencedirect.com/science/article/pii/S0960982220303614},
author = {Rickesh N. Patel and Thomas W. Cronin}}

@ARTICLE{ZhangPolarization,
  author={Zhang, Milin and Wu, Xiaotie and Cui, Nan and Engheta, Nader and Van der Spiegel, Jan},
  journal={Proceedings of the IEEE}, 
  title={Bioinspired Focal-Plane Polarization Image Sensor Design: From Application to Implementation}, 
  year={2014},
  volume={102},
  number={10},
  pages={1435-1449},
  doi={10.1109/JPROC.2014.2347351}}

@ARTICLE{TokudaPolarization,
  author={Tokuda, Takashi and Yamada, Hirofumi and Sasagawa, Kiyotaka and Ohta, Jun},
  journal={IEEE Transactions on Biomedical Circuits and Systems}, 
  title={Polarization-Analyzing CMOS Image Sensor With Monolithically Embedded Polarizer for Microchemistry Systems}, 
  year={2009},
  volume={3},
  number={5},
  pages={259-266},
  doi={10.1109/TBCAS.2009.2022835}}

@ARTICLE{MukulPolarization,
  author={Sarkar, Mukul and San Segundo Bello, David San Segundo and van Hoof, Chris and Theuwissen, Albert},
  journal={IEEE Sensors Journal}, 
  title={Integrated Polarization Analyzing CMOS Image Sensor for Material Classification}, 
  year={2011},
  volume={11},
  number={8},
  pages={1692-1703},
  doi={10.1109/JSEN.2010.2095003}}

@article{HsuPolarization,
author = {Wei-Liang Hsu and Graham Myhre and Kaushik Balakrishnan and Neal Brock and Mohammed Ibn-Elhaj and Stanley Pau},
journal = {Opt. Express},
number = {3},
pages = {3063--3074},
publisher = {OSA},
title = {Full-Stokes imaging polarimeter using an array of elliptical polarizer},
volume = {22},
month = {2},
year = {2014},
url = {http://www.opticsexpress.org/abstract.cfm?URI=oe-22-3-3063},
doi = {10.1364/OE.22.003063},
}

@ARTICLE{Nozaki2017_dvs_temperature_parasitic,
  title     = "Temperature and Parasitic Photocurrent Effects in Dynamic Vision
               Sensors",
  author    = "Nozaki, Y and Delbruck, T",
  journal   = "IEEE Trans. Electron Devices",
  publisher = "ieeexplore.ieee.org",
  volume    =  64,
  number    =  8,
  pages     = "3239--3245",
  month     =  aug,
  year      =  2017,
  url       = "http://dx.doi.org/10.1109/TED.2017.2717848",
  issn      = "0018-9383, 1557-9646",
  doi       = "10.1109/TED.2017.2717848"
}

@INPROCEEDINGS{Graca2021-iisw-unraveling-dvs-noise,
  title           = "Unravelling the paradox of intensity-dependent {DVS} noise",
  booktitle       = "2021 International Image Sensors Workshop ({IISW} 2021)",
  author          = "Graca, Rui and Delbruck, Tobi",
  pages           = "(accepted)",
  year            =  2021,
  url             = "https://imagesensors.org/2021-international-image-sensor-workshop-iisw/",
  conference      = "2021 International Image Sensors Workshop (IISW 2021)",
  location        = "online"
}

@INPROCEEDINGS{Brandli2014-reconstruction,
  title      = "{Real-Time}, {High-Speed} Video Decompression Using a Frame-
                and {Event-Based} {DAVIS} Sensor",
  booktitle  = "Proc. 2014 Intl. Symp. Circuits and Systems ({ISCAS} 2014)",
  author     = "Brandli, Christian and Muller, Lorenz and Delbruck, Tobi",
  pages      = "686--689",
  year       =  2014,
  url        = "http://dx.doi.org/10.1109/ISCAS.2014.6865228",
  address    = "Melbourne, Australia",
  conference = "IEEE International Symposium on Circuits and Systems 2014",
  doi        = "10.1109/ISCAS.2014.6865228"
}

@ARTICLE{Boahen2004-aer-word-serial-tx-design,
  title    = "A burst-mode word-serial address-event link-I: transmitter design",
  author   = "Boahen, K A",
  journal  = "IEEE Trans. Circuits Syst. I Regul. Pap.",
  volume   =  51,
  number   =  7,
  pages    = "1269--1280",
  month    =  jul,
  year     =  2004,
  url      = "http://dx.doi.org/10.1109/TCSI.2004.830703",
  issn     = "1549-8328",
  doi      = "10.1109/TCSI.2004.830703"
}

@INPROCEEDINGS{Berner2007-100-dollar,
  title           = "A 5 {MEPs} \$100 {USB2.0} address-event monitor-sequencer
                     interface",
  booktitle       = "2007 {IEEE} International Symposium on Circuits and
                     Systems",
  author          = "Berner, R and Delbruck, T and Civit-Balcells, A and
                     Linares-Barranco, A",
  publisher       = "IEEE",
  pages           = "2451--2454",
  month           =  may,
  year            =  2007,
  url             = "http://ieeexplore.ieee.org/document/4253172/",
  conference      = "2007 IEEE International Symposium on Circuits and Systems",
  location        = "New Orleans, LA, USA",
  isbn            = "9781424409204, 9781424409211",
  doi             = "10.1109/iscas.2007.378616"
}

@ARTICLE{Gao2014-cup-10to11fps-streak-cam,
  title    = "Single-shot compressed ultrafast photography at one hundred
              billion frames per second",
  author   = "Gao, Liang and Liang, Jinyang and Li, Chiye and Wang, Lihong V",
  journal  = "Nature",
  volume   =  516,
  number   =  7529,
  pages    = "74--77",
  month    =  dec,
  year     =  2014,
  url      = "http://dx.doi.org/10.1038/nature14005",
  language = "en",
  issn     = "0028-0836, 1476-4687",
  pmid     = "25471883",
  doi      = "10.1038/nature14005",
  pmc      = "PMC4270006"
}

@INPROCEEDINGS{Etoh_2011-ultra-hs-ccd,
  title           = "{R57} Progress of Ultra-high-speed Image Sensors with
                     In-situ {CCD} Storage",
  booktitle       = "2011 {INTERNATIONAL} {IMAGE} {SENSOR} {WORKSHOP}",
  author          = "Etoh, T G and Dao, V T S and Nguyen, H D and Fife, K and
                     Kureta, M and Segawa, M and Arai, M and Shinohara, T",
  publisher       = "Intl. Image Sensor Society",
  year            =  2011,
  url             = "https://www.imagesensors.org/Past%20Workshops/2011%20Workshop/2011%20Papers/R57_Etoh_HighSpeedCCD.pdf"
}

@BOOK{Horvath2014-polarization-light-and-vision-in-animals,
  title     = "Polarized Light and Polarization Vision in Animal Sciences",
  editor    = "Horv{\'a}th, G{\'a}bor",
  publisher = "Springer, Berlin, Heidelberg",
  year      =  2014,
  url       = "https://link.springer.com/book/10.1007/978-3-642-54718-8",
  isbn      = "9783642547171, 9783642547188",
  doi       = "10.1007/978-3-642-54718-8"
}

@ARTICLE{Davies2018-loihi,
  title    = "Loihi: A Neuromorphic Manycore Processor with {On-Chip} Learning",
  author   = "Davies, M and Srinivasa, N and Lin, T and Chinya, G and Cao, Y
              and Choday, S H and Dimou, G and Joshi, P and Imam, N and Jain, S
              and Liao, Y and Lin, C and Lines, A and Liu, R and Mathaikutty, D
              and McCoy, S and Paul, A and Tse, J and Venkataramanan, G and
              Weng, Y and Wild, A and Yang, Y and Wang, H",
  journal  = "IEEE Micro",
  volume   =  38,
  number   =  1,
  pages    = "82--99",
  month    =  jan,
  year     =  2018,
  url      = "http://dx.doi.org/10.1109/MM.2018.112130359",
  issn     = "1937-4143",
  doi      = "10.1109/MM.2018.112130359"
}

@ARTICLE{Aimar2018-nullhop,
  title    = "{NullHop}: A Flexible Convolutional Neural Network Accelerator
              Based on Sparse Representations of Feature Maps",
  author   = "Aimar, Alessandro and Mostafa, Hesham and Calabrese, Enrico and
              Rios-Navarro, Antonio and Tapiador-Morales, Ricardo and Lungu,
              Iulia-Alexandra and Milde, Moritz B and Corradi, Federico and
              Linares-Barranco, Alejandro and Liu, Shih-Chii and Delbruck, Tobi",
  journal  = "IEEE Trans Neural Netw Learn Syst",
  month    =  jul,
  year     =  2018,
  url      = "http://dx.doi.org/10.1109/TNNLS.2018.2852335",
  language = "en",
  issn     = "2162-2388, 2162-237X",
  pmid     = "30047912",
  doi      = "10.1109/TNNLS.2018.2852335"
}

@ARTICLE{Chen2017-eyeriss,
  title    = "Eyeriss: An {Energy-Efficient} Reconfigurable Accelerator for
              Deep Convolutional Neural Networks",
  author   = "Chen, Yu-Hsin and Krishna, Tushar and Emer, Joel S and Sze,
              Vivienne",
  journal  = "IEEE J. Solid-State Circuits",
  volume   =  52,
  number   =  1,
  pages    = "127--138",
  month    =  jan,
  year     =  2017,
  url      = "http://dx.doi.org/10.1109/JSSC.2016.2616357",
  issn     = "0018-9200, 1558-173X",
  doi      = "10.1109/JSSC.2016.2616357"
}

%TC:endignore
\end{document}